%% file: main.tex
\documentclass[runningheads]{llncs}

\usepackage{eccv}

\usepackage{eccvabbrv}

\usepackage{graphicx}
\usepackage{booktabs}
\usepackage{multirow}

\usepackage[accsupp]{axessibility}  %

\usepackage{hyperref}

\usepackage{orcidlink}

\newcommand{\methodname}{BAT3R\xspace}

\begin{document}

\title{Bootstrapping Articulated 3D Reconstruction from 2D Image Collections}

\titlerunning{Bootstrapping Articulated 3D Reconstruction from 2D Image Collections}

\author{Jakub Zadrozny \and
Oisin Mac Aodha~\orcidlink{0000-0002-5787-5073} \and
Hakan Bilen~\orcidlink{0000-0002-6947-6918}}

\authorrunning{J.~Zadrozny et al.}

\institute{University of Edinburgh\\
\href{https://jakubzadrozny.github.io/bat3r}{\nolinkurl{jakubzadrozny.github.io/bat3r}}}

\maketitle

\begin{abstract}
3D reconstruction of  articulated objects from a single image is challenging because large training datasets with paired image and 3D supervision are difficult to obtain. 
Recent point map–based methods achieve strong performance but rely on synthetic datasets rendered from manually created articulated 3D assets with carefully curated pose distributions. While camera viewpoints can be easily sampled, generating realistic object articulations remains costly and labor-intensive. We propose a training framework that reduces this requirement by leveraging unannotated 2D images collections with only a single rigged canonical mesh per category.
Starting from a weak 3D shape predictor trained on canonical-pose renders, we iteratively estimate object articulation and camera pose by fitting the mesh to predicted point maps. The recovered articulations and viewpoints are then used to render updated synthetic training data, progressively improving the predictor. 
Despite using substantially weaker 3D supervision, our models achieve performance comparable with DualPM, which requires manually curated articulated training datasets.
\end{abstract}

\section{Introduction}
\label{sec:intro}
Estimating the 3D shape of an object from a single image is an inherently ill-posed task in computer vision, as multiple 3D structures can correspond to the same 2D observation. 
The challenge becomes even more pronounced in the presence of occlusions, whether self-occlusions or those caused by other objects in the scene, which leave only partial information about the object's geometry.
Addressing this problem therefore requires reasoning about the complete 3D structure from limited visual evidence.
This ability to infer and understand such complex 3D shapes is fundamental for next-generation embodied systems that must perceive, reason about, and interact with objects in the physical world.

Previous works addressing this problem typically rely on incorporating priors to constrain the otherwise ambiguous 3D reconstruction space, ensuring that predicted shapes are both plausible and consistent with the input image.
Such priors have been introduced in several ways: through morphable 3D shape templates~\cite{loper2023smpl,zuffi20173d}, learned from 2D image collections~\cite{aygun2024saor,wu2023magicpony,li2024learning} or by encoding shape knowledge into neural networks with supervised 3D data~\cite{kaye2025dualpm,xiang2025structured}. 
Among these approaches, supervised models trained on paired input images and corresponding 3D data (\eg meshes) have demonstrated particularly strong performance~\cite{chen2025sam}. 

However, like most supervised approaches, the success of these methods is heavily reliant on the quality and diversity of available training data. 
While recent efforts have expanded the availability of synthetic 3D assets through large-scale dataset curation~\cite{deitke2023objaverse,deitke2023objaverseXL,li20214dcomplete},
these collections still suffer from limitations in realism and diversity.
Moreover, the manual curation required to build larger, high-quality datasets is prohibitively expensive.
This issue is especially pronounced for object-centric 3D reconstruction of deformable and articulated objects  (\eg animals), which require substantial amounts of diverse 3D data for effective supervised training. 
Outside of human-centric domains~\cite{cai2024playing,yang2023synbody,black2023bedlam,tesch2025bedlam2}, there remains a scarcity of 3D datasets for other complex articulated objects.

\begin{figure}[t]
\centering
\includegraphics[width=\textwidth]{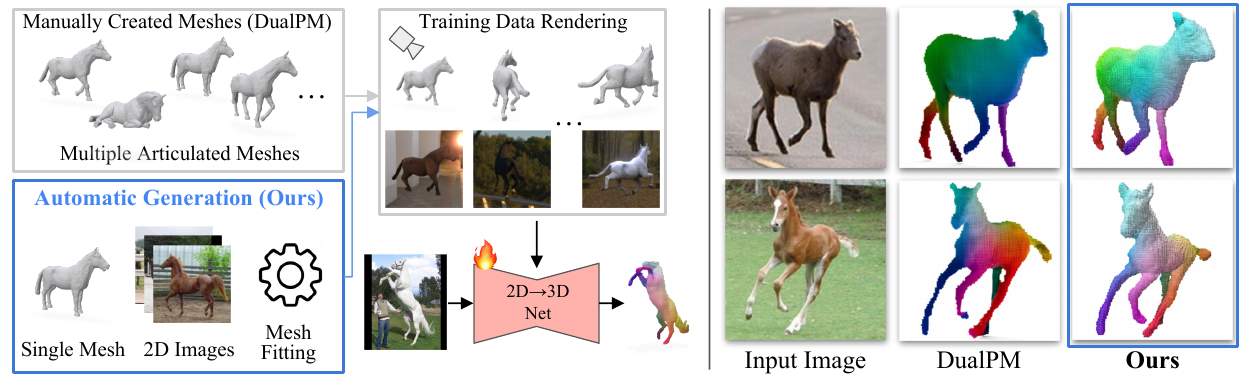}
\vspace{-15pt}
\caption{
While effective, supervised approaches for 3D shape estimation from 2D images rely on the availability of large manually crafted and curated 3D shape datasets depicting objects of interest in a wide variety of poses (top left). 
We introduce a new approach that only requires a single mesh per category and an image collection as input (bottom left). 
From this, we are able to automatically generate paired training data, \ie 2D images and corresponding 3D point clouds, that can be used to train supervised reconstruction methods. 
Our 3D prediction results are comparable to methods that are trained on richer manually created 3D articulated data, \eg DualPM~\cite{kaye2025dualpm} (right). 
}
\label{fig:motivation}
\end{figure}

While obtaining large-scale articulated 3D data for a specific object category is challenging, as discussed above, single meshes depicting the category in a rest (\ie canonical or neutral) pose are often readily available. %
Motivated by this observation, we ask whether it is possible to train a model to estimate the 3D shape of an object from a single image given only a single 3D model of that category and a set of 2D images at training time? %
In doing so, we would significantly decrease the amount of manual curation required to generate such datasets and provide a scalable pathway to training models for categories that have limited 3D data available. 

At first glance, this appears infeasible as a single mesh cannot capture the wide range of pose and shape variation observed in real-world images. 
Nevertheless, we observe that training from rendered images obtained from a single 3D mesh can provide a strong initialization for basic shape and camera viewpoint estimation (see \cref{fig:motivation} for an overview). 
Building on this insight, we propose an iterative data generation procedure that progressively augments the training set with predicted paired 2D and 3D training instances that exhibit progressively more complex poses.  
In doing so, we demonstrate that it is possible to train the recent state-of-the-art DualPM~\cite{kaye2025dualpm} 3D shape estimation method on significantly weaker 3D data yet still obtain comparable performance on challenging quadruped animal categories.  

We make the following contributions: 
(i)  We propose a new approach, Bootstrapping ArTiculated 3D Reconstruction (\methodname),  for automatically generating paired 2D and 3D training data that can be used to train category-centric 2D to 3D shape prediction models while foregoing the need for expensive manually created 3D articulated training data. 
(ii) Despite being trained on significantly weaker 3D data, we demonstrate that our approach results in strong reconstruction performance. 
Furthermore, we show that the iterative  application of our method improves performance, closing the gap between our approach and fully supervised baselines.

\section{Related work}
\label{sec:rel_work}

There is a large body of work in computer vision that attempts to reconstruct a 3D representation of a scene given a set of images depicting the scene, \eg \cite{schoenberger2016sfm,murez2020atlas,mildenhall2021nerf,kerbl20233d,wang2024dust3r}.  
In this work, our focus is on a different type of 3D reconstruction task, that of  reconstructing deformable objects from unstructured image collections. 

\subsubsection*{Unsupervised reconstruction.}
There is a family of unsupervised methods that have been proposed that can estimate the 3D shape of a \emph{rigid} object category given a collection of images~\cite{vicente2014reconstructing,kanazawa2018learning,goel2020shape,li2020self,monnier2022share}. 
Unlike conventional structure from motion methods that assume the provided images are from different viewpoints of same object instance or scene at similar points in time, these methods instead take an unstructured set of images as input, \ie a set of images depicting the same object category, but not necessarily the same exact instance. 
The methods are unsupervised in the sense that no ground-truth 3D supervision is available, but they instead make use of auxiliary 2D supervision such as foreground segmentation masks, semantic part segmentation masks, object keypoints, camera poses, \etc. 
Successive methods have managed to remove some of the supervision signal required either partially (\eg  by removing the need for keypoint annotations~\cite{goel2020shape}) or completely~\cite{monnier2022share}. 
Due the rigid object assumptions made by these methods, they cannot estimate plausible 3D shapes for articulated/deformable objects. 

Techniques have been proposed to address the more challenging non-rigid case whereby objects may deform or articulate. 
Given a collection of images and a skeleton, LASSIE~\cite{yao2022lassie} introduced a test-time optimization approach to reconstruct articulated categories. 
Later this was extended such that the skeleton could also be estimated from the images~\cite{yao2023hi}. 
An alternative approach is to train a neural network to directly predict the articulated 3D shape. 
Skeleton-based priors have also been demonstrated to be effective in this setting as in MagicPony~\cite{wu2023magicpony} or 3D-Fauna~\cite{li2024learning}, but other priors such as motion cues obtained from video sequences, as used in DOVE~\cite{wu2023dove}, are also effective. 
SAOR\cite{aygun2024saor} does not require any 3D or video supervision, instead relying on the assumption that objects can be decomposed into a set of articulated parts which are learned during training. 
While impressive, the results of these unsupervised methods are still inferior to techniques that are trained with full 3D supervision.

\subsubsection*{Supervised reconstruction.} 
The above unsupervised methods broadly do not make use of any detailed 3D shape information during training. 
However, when available, this can impose a strong regularization signal ensuring that model predictions stay on the manifold of plausible 3D shapes.  
One approach is to align multiple 3D meshes of different instances of the same object category into a common model which captures the shape variation within the category. 
This has been shown to be highly effective for humans~\cite{loper2023smpl}, quadrupeds~\cite{zuffi20173d} more generally, and specific species such as dogs~\cite{ruegg2023bite,rueegg2023barc} and horses~\cite{zuffi2024varen}. 
However, the 3D alignment step is non-trivial and limits this approach to settings where a large number of 3D meshes containing instances of the same topology are available. 

When large quantities of paired image and 3D data are available, it is possible to directly apply supervised learning techniques to learn the mapping from images to 3D, \eg~\cite{zou2024triplane}. 
Additionally, priors learned from deep generative (\eg diffusion-based) models have also been shown to be effective at assisting with 3D extraction~\cite{liu2023zero,long2024wonder3d,xiang2025structured}. 
These supervised approaches require training data, but advances in rendering and the availability of large 3D asset collections have enabled the creation of paired datasets in the case of a limited number of deformable categories such as animals, \eg~\cite{xu2023animal3d,niewiadomski2025generative,jakab2024farm3d}. 
However, the generation of such datasets relies on the availability of 3D assets in diverse poses.

Inspired by point-based representations developed in the context of 3D reconstruction networks~\cite{wang2024dust3r} and canonical surface mapping~\cite{kulkarni2019canonical,kulkarni2020articulation}, DualPM~\cite{kaye2025dualpm} introduced a supervised approach for viewpoint invariant point map prediction for deformable object categories.  
It leverages large synthetically rendered images and 3D point clouds to train a supervised model that can predict both the 3D position of visible and occluded parts of the object via a multi-layered output representation. 
DualPM produces state-of-the-art reconstruction results for quadrupeds. 
However, like the supervised methods previously discussed, it requires 3D training data depicting objects in diverse poses. 
In this work, we propose a new approach for automatic dataset generation that overcomes this requirement for diverse data. 
Specifically, we show that it is possible to generate rich training data via an iterative procedure that only requires a single rigged canonically posed mesh as input.

\section{Method}
\label{sec:method}

Given a collection of RGB images $\{I_n\}_{n=1}^N$ depicting different instances of an articulated object category (\eg horses), our goal is to train a single-image predictor $\Phi$ that reconstructs the object's 3D geometry.
At training time, we assume we have access to a single canonical mesh $M^c$ representing the object in its rest pose. We also assume that $M^c$ is animation-ready, \ie rigged and skinned with an articulation skeleton. Such meshes can be created manually or obtained using recent automatic rigging methods~\cite{zhang2025one,song2025puppeteer}. 
Importantly, we do not require additional instances of this mesh in different articulated poses.

For the image collection, we assume foreground segmentation masks are available, which can be obtained using off-the-shelf segmentation methods~\cite{kirillov2023segment,ravi2024sam}. The images require neither camera pose annotations nor 3D supervision. We also do not assume prior knowledge of the distribution of valid object articulations, and only require a coarse prior over camera viewpoints. 
Both object articulations and camera poses are estimated automatically during training.
We use the DualPM point-map representation~\cite{kaye2025dualpm} for the single-image predictor, which we summarize next.

\subsection{Preliminaries: Dual Point Maps (DualPM)}
\label{sec:dualpm}

Our main contribution is a framework for automatically generating paired 2D--3D training data for articulated image-to-3D reconstruction. For the reconstruction model, we build on DualPM~\cite{kaye2025dualpm}, which predicts two coupled, layered 3D point maps from a single image.

Given an input image $I$, for each foreground pixel $u$, the network predicts $K$ intersection points between a ray passing through pixel $u$ and the object surface (capturing both visible and occluded parts). Each intersection point at layer $k$ is represented by a posed point $P_k(u) \in \mathbb{R}^3$ and a canonical point $Q_k(u) \in \mathbb{R}^3$. The posed point is expressed in the camera coordinate frame, while the canonical point is expressed in the object's rest-pose coordinate system. Together, $(P_k(u), Q_k(u))$ describe the same physical surface location in two coordinate frames. The posed map therefore provides a 3D reconstruction of the object in the camera coordinate frame, while the canonical map establishes dense correspondences mapping canonical points to posed points. This explicit canonical-to-posed mapping is the core mechanism driving our mesh fitting stage.

DualPM~\cite{kaye2025dualpm} is a fully supervised approach and it is trained using synthetically rendered images depicting articulated meshes where ground-truth posed and canonical coordinates are available for each surface point (both visible and occluded). 
We let $\hat{P}_k(u)$ and $\hat{Q}_k(u)$ denote the predicted posed and canonical points, respectively, and $F$ denote the set of masked object (foreground) pixels.
The posed point map is trained using the confidence-weighted regression loss: 
\begin{equation}
\mathcal{L}_P =
\frac{1}{K|F|}
\sum_{u \in F}
\sum_{1 \leq k \leq K}
c_{P_k}(u)\, \| \hat{P}_k(u) - P_k(u) \|_2^2
- \alpha \log c_{P_k}(u),
\end{equation}
where ${P}_k(u)$ and ${Q}_k(u)$ denote the ground truth posed and canonical points, $c_{P_k}(u)$ denotes a predicted confidence for the posed point estimate. 
The total training objective is  $\mathcal{L}_{DualPM} = \mathcal{L}_{P} + \mathcal{L}_{Q}$, where an analogous loss to $\mathcal{L}_P$ is used for $\mathcal{L}_Q$.

\begin{figure}[t]
\centering
\includegraphics[width=\textwidth]{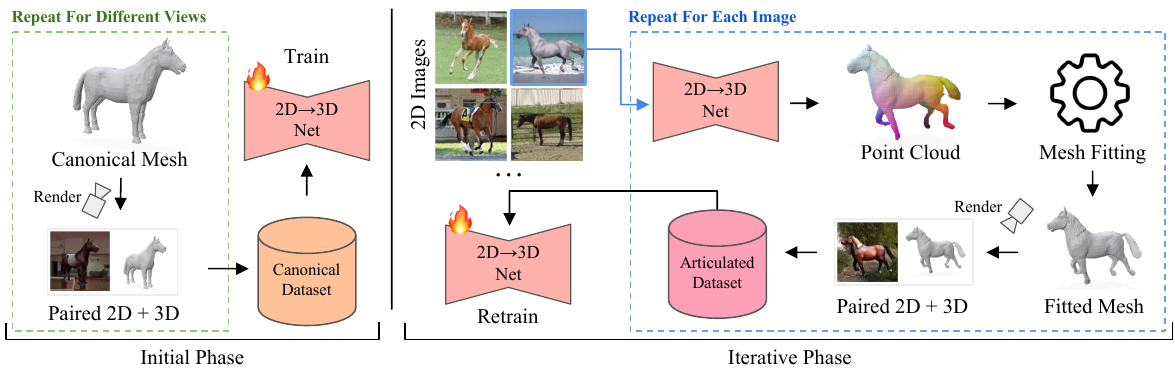}
\vspace{-10pt}
\caption{
Overview of our bootstrapping approach \methodname. Left: We render the canonical rigged mesh from sampled camera viewpoints to create an initial paired 2D--3D training set, which is used to train the initial 2D to 3D  predictor $\Phi_0$. Right: Given an image collection, the current predictor $\Phi_i$ predicts posed and canonical point maps. We fit the canonical mesh to these predictions, recover an articulated mesh and camera viewpoint, and render the result to obtain new synthetic training pairs with exact point-map supervision. The predictor is then retrained on this generated dataset, yielding $\Phi_{i+1}$. The process is repeated for multiple iterations.
}
\label{fig:method_overview}
\end{figure}

\subsection{\methodname: Bootstrapping ArTiculated 3D Reconstruction}
The original DualPM framework requires a training collection of 2D images  depicting the object of interest in various articulated poses, paired with corresponding ground-truth layered 3D point maps with canonical correspondences.
In practice, this paired data is generated synthetically via layered rasterization of articulated 3D meshes.
However, acquiring such high-quality datasets is highly non-trivial as creating  realistic articulation data for rigged meshes is a challenging, time-consuming, and expensive process that typically demands the expertise of professional 3D artists. 
Furthermore, achieving comprehensive coverage of natural articulations is critical, as the predictive performance and generalization capabilities of the learned model are inherently limited by the pose and articulation diversity present within the training data. 
We propose an alternative solution that requires significantly less 3D supervision. 
Our method alternates between two steps: fitting the canonical mesh to predictions on the 2D image collection, and using the fitted meshes to render new paired 2D–3D training data.

We assume we have access to an initial, potentially weak, 3D predictor $\Phi_i$, which can be initialized by training the predictor on image and 3D data rendered from a single canonical mesh (\ie depicting a single pose).  
The high-level idea behind our approach is that we can use the initial predictor to generate plausible predictions for the supplied image collection and then make use of mesh alignment methods to fit the canonical mesh to the predicted point clouds to obtain realistic object poses that can be used to generate improved ground-truth 3D training data. %
Below we detail a single iteration of our proposed refinement procedure, designed to yield an improved predictor, $\Phi_{i+1}$. 
An illustration of our method, \methodname, is depicted in \cref{fig:method_overview}. 

We first employ the current model $\Phi_i$ to infer the dual point clouds (derived from the dual point maps), \ie the posed point cloud $P$ and the corresponding canonical point cloud $Q$, for each image $I$ in the training collection. 
Subsequently, we optimize for the best articulation of the provided canonical 3D mesh to the predicted posed point cloud for each instance. %
This fitting process is facilitated by the semantic correspondences provided by the dual point cloud representation (see \cref{sec:dualpm}). 
Although the geometric fidelity of the predicted point clouds may be imperfect (particularly during early stages of the refinement), their primary function is merely to guide the mesh articulation toward the target pose depicted in the input image. 
A key observation motivating our approach is that exact ground-truth 3D annotations for the input images are not required to iteratively improve the predictor.
Instead, we rely on articulating the canonical mesh to fit the imperfect predictions, which inherently provides a fully accurate and self-consistent 3D structure for supervision. 
Furthermore, as a natural by-product of our mesh alignment process, we also recover the camera poses corresponding to the input images. %
Finally, these articulated meshes and estimated camera poses are utilized to render a novel, self-annotated training set, which serves as the supervisory signal to train the subsequent model iteration, $\Phi_{i+1}$.

\subsubsection{Initialization.} In the most general case, we begin the process assuming no knowledge about the distribution of plausible 3D articulations and simply train a naïve initial model to serve as $\Phi_0$, \ie $\Phi_i$ for the first iteration where $i = 0$.
To this end, we automatically create a dataset of 2D renders of the canonical mesh $M^c$ (\ie a 3D mesh of the object in a rest pose) from many randomly sampled camera viewpoints. 
The distribution used to sample camera viewpoints should in general resemble the distribution that is expected to be encountered at test time. 
To complete the rendering pipeline we require an object texture, which can be hand-crafted or AI-generated, and a environment map which we sample from a collection of publicly available assets. %
This dataset of images of the object in the canonical pose from various  viewpoints, with known 3D per-image shape,  serves as our initial paired training data.

\subsubsection{Inference and mesh fitting.}
Our mesh fitting algorithm employs a gradient-based optimization loop to articulate and align the canonical 3D mesh to the predicted dual point clouds. 
The dual point map representation lends itself particularly well to this purpose by establishing explicit dense correspondences between the posed and the canonical states.
Specifically, let $V^c = \{v^c_j\}$ denote the vertices of the canonical mesh $M^c$, and let $P = \{p_i\}$ and $Q = \{q_i\}$ represent the predicted (using $\Phi_i$) posed and canonical point clouds for a specific image $I$, respectively. 
First, each point indexed by $i$ in the dual point map is assigned to the mesh vertex that minimizes the distances in the canonical space, \ie $j^*(i) = \arg\min_j \|q_i - v^c_j\|_2$.
Intuitively, during the fitting process, the corresponding predicted posed point $p_i$ acts as a target attractor pulling on the articulated mesh vertex.

Furthermore, when fitting the mesh to point clouds predicted from in-the-wild images, the template mesh $M^c$ will inherently exhibit anatomical discrepancies compared to the observed instance. To prevent the optimization from overfitting to these instance-specific structural variations, we subsample the predicted dual point clouds to a sparse set of 200 control points.
To ensure these control points are sampled uniformly along the surface of the object, we employ a topology-aware farthest point sampling strategy. We define a surface-grounded distance metric $D(q_n, q_m)$ between any two predicted canonical points $q_n, q_m \in Q$. Let $j^*(n)$ denote the index of the mesh vertex assigned to $q_n$, and let $d_{geo}(\cdot, \cdot)$ represent the geodesic distance along the canonical mesh surface. We formulate the distance as the path length bridging the two points via the mesh surface:
\begin{equation}
    D(q_n, q_m) = \|q_n - v^c_{j^*(n)}\|_2 + d_{geo}\left(v^c_{j^*(n)}, v^c_{j^*(m)}\right) + \|v^c_{j^*(m)} - q_m\|_2
\end{equation}
For ease of notation, in all subsequent sections, we assume $P$ and $Q$ refer directly to these subsampled control point clouds.

The optimization process begins by determining an optimal global rigid transformation $T^g$ (comprising scaling, rotation, and translation) using the Kabsch algorithm~\cite{kabsch1976solution}, to align the canonical mesh (defined in the rest-pose coordinate frame) with the predicted posed point cloud $P$ (residing in the camera space). 
Following this global alignment, we optimize the articulation parameters $\theta$, defined as the excess joint rotations relative to the rest pose, alongside per-bone (scalar) scale parameters $S$ to account for instance-specific anatomical proportions (\eg variations in limb length).
The state of the posed mesh is thus a function of the articulation, bone scales, and global rigid transform, yielding the posed vertices $V^P(\theta, S, T^g) = \{v^p_j(\theta, S, T^g)\}$.

To guide this process, we define a loss function consisting of a data term and geometric regularization constraints. 
The data term, $\mathcal{L}_{data}$, measures the discrepancy between each predicted posed point $p_i$ and its assigned articulated mesh vertex $v^p_{j^*(i)}$: 
\begin{equation}
\mathcal{L}_{data} = \sum_{p_i \in P} \mathcal{H}_\delta \left( \lVert p_i - v^p_{j^*(i)}(\theta, S, T^g) \rVert_2 \right).
\end{equation}
To mitigate the influence of outliers and to accommodate natural deviations between the predicted point cloud and the specific instance's true shape, we employ a robust Huber loss $\mathcal{H}_\delta$ with $\delta=0.1$ for this distance.

Our objective is not only to overfit the point cloud, but rather to find a close, reasonable alignment while preserving the intrinsic structural properties of the fitted mesh, which ensures the 3D predictor $\Phi$ learns a geometrically correct prior. 
To this end, we incorporate a set of regularization losses to address overfitting. 
First, inspired by physical energy minimization principles, we penalize the mean squared magnitude of the excess joint rotations ($\|\theta\|_2^2$), which encourages smooth and gradual curvatures. 
Second, we introduce penalties against deviations in (local) volume $\mathcal{L}_{vol}$ and edge lengths $\mathcal{L}_{edge}$, between the canonical and articulated states, thereby preventing unrealistic stretching, shearing, or compression. %
Third, to ensure anatomical consistency, we regularize the bone scales via $\mathcal{L}_{scale}$, which enforces left-right symmetry and penalizes extreme deviations from the canonical proportions.
Finally, when presented with challenging articulations, the 3D predictor tends to output noisy point clouds, especially early in the refinement process, which in turn results in self-intersections of the fitted meshes. 
To mitigate this effect, we apply a self-repulsion loss term, $\mathcal{L}_{rep}$, which explicitly penalizes configurations where vertices separated by a large geodesic distance on the mesh surface collapse to a small Euclidean distance in the articulated state, effectively preventing self-intersections. 
Crucially, to prevent these structural penalties from opposing the valid anatomical size variations captured by $S$, the geometric losses ($\mathcal{L}_{vol}, \mathcal{L}_{edge}, \mathcal{L}_{rep}$) are computed on the \textit{unscaled} articulated mesh.
We provide more details in the supplementary material. 

The complete objective function is formulated as:
\begin{equation}
\label{eq:loss}
\mathcal{L} = \mathcal{L}_{data} + \lambda_{angle} \lVert \theta \rVert_2^2 + \lambda_{vol} \mathcal{L}_{vol} + \lambda_{edge} \mathcal{L}_{edge} + \lambda_{scale} \mathcal{L}_{scale} + \lambda_{rep} \mathcal{L}_{rep}.
\end{equation}
The total loss $\mathcal{L}$ is jointly minimized using gradient descent with respect to the excess joint rotations $\theta$, bone scales $S$, and the global rigid transform $T^g$, as articulating the mesh necessitates corrections to the global alignment.
Importantly, the bone scales $S$ act solely as auxiliary variables to absorb instance-specific anatomical variations for a more accurate data fit. Upon convergence, $S$ is discarded, and we extract the unscaled articulated mesh to serve as the geometrically consistent 3D training target for the next iteration.
Note that our regularization terms, apart from the angular penalty, which we assign a low weight $\lambda_{angle}$, do not hinder intricate yet natural deformations, as physically plausible articulations generally do not necessitate excessive stretching, compression, volume fluctuations, or self-intersections.

We empirically observe that our approach proves more effective at extracting accurate poses from lateral views compared to frontal and rear perspectives. To address this imbalance and boost performance on the challenging views, we employ a \textit{viewpoint augmentation} strategy. Specifically, to form the training set for the next iteration, we render the fitted meshes from multiple different novel viewpoints (see supplement for details).

\section{Experiments}
\label{sec:exps}

\subsection{Experimental setup}
\label{sec:setup}
\noindent{\textbf{Datasets.}}
For our synthetic data evaluation, we strictly follow the training splits provided by DualPM~\cite{kaye2025dualpm} for the horse, cow, and sheep categories. Crucially, we do not access the ground-truth articulations and camera poses. 
Instead, we recover the articulations and camera poses using our refinement procedure.
For our real-world evaluation, we collect purely 2D in-the-wild images. For horses, we utilize ${\sim}11,000$ real images from the MagicPony~\cite{wu2023magicpony} dataset. 
For chimpanzees, we assemble a training collection of ${\sim}19,000$ images from AP-10K~\cite{ap10k} and OpenApePose~\cite{desai2023openapepose} (restricted to chimpanzees). 
Finally, for elephants, we assemble a training collection of ${\sim}21,000$ images comprising the \textit{Wild Elephant Dataset} from Kaggle and frames extracted from YouTube videos using the Animal-in-Motion~\cite{zhao2026webscale} pipeline.

\noindent{\textbf{3D assets and rendering.}}
Our approach requires a canonical 3D mesh. For the horse, cow, and sheep categories, we utilize the exact canonical assets provided by DualPM. For chimps and elephants, we acquire publicly available meshes from the internet. While we utilize manually rigged meshes in this work, our pipeline is agnostic to the rigging source and can be paired with modern auto-rigging solutions \cite{liu2025riganything,zhang2025one}.

\noindent{\textbf{Evaluation protocol.}}
Our evaluation spans both strict baseline comparisons and wider generalization tests. To this end, we utilize three quantitative test splits for the horse category: (1) \textit{Animodel-Points}~\cite{jakab2024farm3d, kaye2025dualpm}: the exact test split defined by DualPM; (2) \textit{Horse-Robust}: a comprehensive, large-scale dataset of $15,000$ samples (compared to ${\sim}500$ in Animodel-Points), where only the male horse template is available at training time, but the female template is used for evaluation; (3) \textit{Horse-Mixed}: a standard in-distribution split ($18,000$ samples) where models are both trained and evaluated on a mix of male and female templates. We utilize this split for controlled ablation studies.

\noindent{\bf Evaluation metrics.} 
Following the Animodel-Points benchmark protocol~\cite{jakab2024farm3d, kaye2025dualpm}, we evaluate the geometric accuracy of our predicted point clouds using the Root-Mean-Square (RMS) bi-directional Chamfer Distance (CD), reported in centimeters. 
We evaluate under both full rigid alignment (using Iterative Closest Point (ICP)) and model-view alignment (restricted to scale and translation), thereby penalizing inaccurate camera pose predictions. 
In the ablation studies, we report with more granularity, \ie the mean CD (mCD), the outlier ratio denoted as \% (CD $>$ 5cm) which measures the fraction of test samples exceeding a 5cm error threshold, and the angular camera rotation error $\Delta R$ ($^\circ$).

\noindent{\textbf{Baselines.}}
We compare against category-specific models~\cite{kulkarni2020articulation,wu2023magicpony,jakab2024farm3d,li2024learning,xiang2025structured}, including the state-of-the-art DualPM~\cite{kaye2025dualpm}, which is trained with full 3D supervision. For real-world categories where 3D supervision is unavailable, we compare qualitatively against recent zero-shot 3D foundation models, namely SAM-3D~\cite{chen2025sam} and Trellis.2~\cite{xiang2026native}, which are evaluated in a zero-shot manner.

\noindent{\textbf{Implementation summary.}}
We employ the DualPM architecture as our point map predictor. When training on synthetic images, we disable bone scaling, as these datasets lack the individual anatomical variation present in real-world samples. Furthermore, we observed that the original DualPM training regime becomes unstable when supervised by the mesh articulations fitted to noisy initial predictions, which inherently contain extreme outliers despite geometric regularization (particularly in the case of real-world images).
To mitigate this instability, we adopt a modified optimization regime tailored for noisy, self-supervised targets (see the supplementary material for details).
We typically run our refinement loop for four iterations (see \cref{sec:ablations} for analysis). Further implementation details, including texture generation, camera pose distributions, and mesh fitting parameters, are provided in the supplementary material.

\begin{table*}[t]
\centering
\caption{Quantitative evaluation on the Animodel-Points dataset. We report the Root Mean Square Error (RMS) of the bi-directional Chamfer Distance in centimeters ($\downarrow$). Our methods in this comparison (`Canonical' and `Ours (synt)') are trained on the synthetic images from DualPM following original splits. We report both the results with and without rotation alignment. Best results in \textbf{bold}, 2nd best \underline{underlined}.
}
\resizebox{\textwidth}{!}{
\begin{tabular}{l|ccc|ccc}
\toprule
\multirow{2}{*}{Method} & \multicolumn{3}{c|}{RMS Chamfer Distance (cm) $\downarrow$} & \multicolumn{3}{c}{Model-view RMS Chamfer Dist.~(cm) $\downarrow$} \\
\cmidrule(lr){2-4} \cmidrule(lr){5-7}
& Horse & Cow & Sheep & Horse & Cow & Sheep \\
\midrule
A-CSM~\cite{kulkarni2020articulation} & $11.75 \pm 3.83$ & $9.52 \pm 2.41$ & $9.24 \pm 2.40$ & $38.13 \pm 13.89$ & $33.51 \pm 11.52$ & $29.04 \pm 9.35$ \\
MagicPony~\cite{wu2023magicpony} & $11.19 \pm 3.08$ & $10.29 \pm 2.08$ & -- & $20.82 \pm 13.04$ & $25.39 \pm 13.43$ & -- \\
Farm3D~\cite{jakab2024farm3d} & $11.34 \pm 3.22$ & $9.63 \pm 2.02$ & $11.01 \pm 1.87$ & $29.52 \pm 15.73$  & $21.34 \pm 12.85$ & $21.52 \pm 9.84$ \\
3D-Fauna~\cite{li2024learning} & $11.86 \pm 3.03$ & $10.54 \pm 2.26$ & $9.61 \pm 2.15$ & $15.70 \pm 6.82$ & $14.08 \pm 4.20$ & $12.24 \pm 3.17$ \\
Trellis~\cite{xiang2025structured} & $6.93 \pm 4.13$ & $6.80 \pm 3.24$ & $5.91 \pm 2.93$ & $36.82 \pm 16.02$ & $26.54 \pm 13.14$ & $26.56 \pm 12.06$ \\
DualPM~\cite{kaye2025dualpm} & $\mathbf{4.30 \pm 1.50}$ & $\mathbf{3.18 \pm 1.06}$ & $\mathbf{3.30 \pm 1.20}$ & $\mathbf{5.49 \pm 1.75}$ & $\mathbf{4.03 \pm 2.11}$ & $\mathbf{4.22 \pm 2.18}$ \\
\hline
Canonical & $7.88 \pm 3.01$ & $6.66 \pm 2.31$ & $6.62 \pm 2.33$ & $9.87 \pm 4.55$ & $7.85 \pm 2.87$ & $7.70 \pm 2.66$ \\
\textbf{Ours (synt)} & \underline{$5.65 \pm 1.57$} & \underline{$4.17 \pm 1.15$} & \underline{$4.35 \pm 1.26$} & \underline{$7.73 \pm 2.06$} & \underline{$5.30 \pm 2.42$} & \underline{$5.04 \pm 1.71$} \\
\midrule
\end{tabular}
}
\label{tab:main_res}
\end{table*}

\begin{table*}[t]
\centering
\caption{Quantitative evaluation on the Horse-Robust test split. We report the Root Mean Square Error of the bi-directional Chamfer Distance (RMS CD) and Model-view (MV) RMS CD. Both variants of our approach outperform large-scale foundation models despite using orders of magnitude less 3D supervision. 
Trellis.2 relies purely on best-effort post-hoc rotation alignment which occasionally fails, inflating its error. Best results in \textbf{bold}, 2nd best \underline{underlined}.
}
\begin{tabular}{l | ccc | ccc}
\toprule
Metric (cm) $\downarrow$ & Trellis.2 & SAM-3D & DualPM & Canonical & \textbf{Ours (synt)} & \textbf{Ours (real)} \\
\midrule
RMS CD    & 8.92 & 6.42 & \textbf{3.84} & 7.84 & \underline{5.10} & 6.01 \\
MV RMS CD & 40.03 & 8.71 & \textbf{5.42} & 10.83 & \underline{7.92} & 8.37 \\
\bottomrule
\end{tabular}
\label{tab:horse-robust}
\end{table*}

\subsection{Results}

\noindent{\bf Training on synthetic images.}
\label{sec:results-synthetic}
In \cref{tab:main_res} we quantitatively compare our approach to existing single image 3D object reconstruction methods. 
We compare to both unsupervised and supervised methods, across the three object categories (horse, cow, and sheep) evaluated in DualPM.
DualPM reports state-of-the-art performance on this task, and serves as a fully supervised upper bound for our approach. 
We observe that the model trained only on images rendered from the canonical mesh (`Canonical') is surprisingly effective, specifically on the comparatively easier sheep category which exhibits less articulation than horses. 
In general, our method, trained on the exact synthetic image splits provided by DualPM, approaches its performance while also outperforming the other baselines tested.
See the supplementary material for a qualitative comparison with 3D-Fauna~\cite{li2024learning} and DualPM~\cite{kaye2025dualpm} on the \textit{cow} and \textit{sheep} categories.

\noindent{\bf Training on real images.}
Here we demonstrate the capabilities of our approach trained exclusively on real-world in-the-wild image collections.
In \cref{tab:horse-robust} we compare the performance of our model trained on real-world images (`Ours (real)') to DualPM, our model trained only on images rendered from the canonical mesh (`Canonical'), our model trained on synthetic images (`Ours (synt)'), and large image-to-3D supervised  foundation models~\cite{chen2025sam,xiang2026native} on the wide \textit{Horse-Robust} evaluation set. 
Our approach successfully leverages the pose diversity contained in real-world images to improve performance from the `Canonical' initialization point close to the version trained on synthetic images (which contain far less inherent real-world diversity and hence the signal is easier to extract). Importantly, `Ours (real)' outperforms large-scale supervised models, despite using orders of magnitude less 3D supervision.

\begin{figure}[h]
\centering
\includegraphics[width=\textwidth]{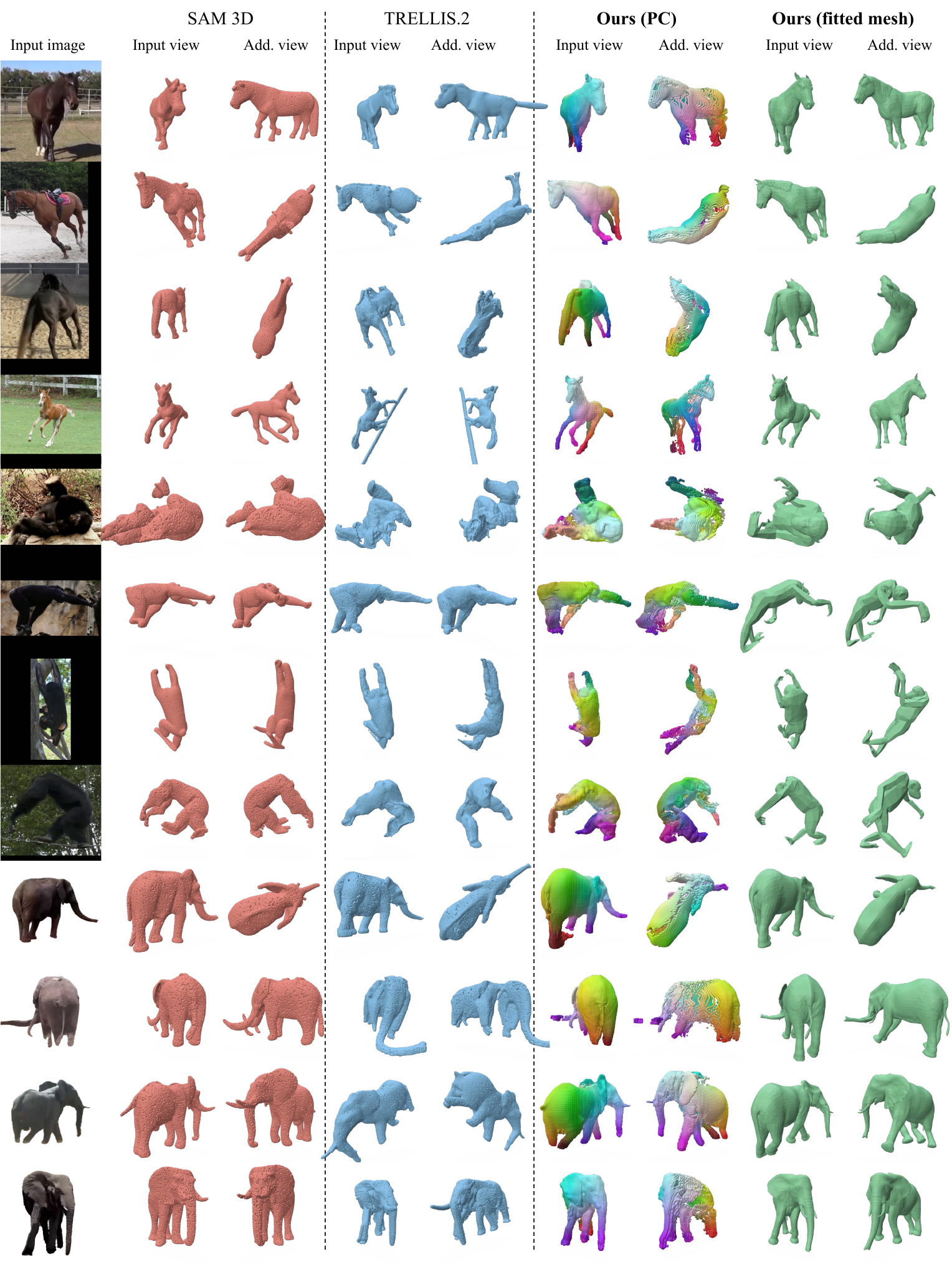}
\caption{Qualitative comparison on real-world images across three challenging categories: horse, chimpanzee, and elephant. 
Trellis.2~\cite{xiang2026native} struggles with accurate real-world conditioning and relies on post-hoc rigid ICP alignment of their predicted 3D shape to a reference predicted point cloud (here we use 'Ours').
SAM-3D~\cite{chen2025sam} exhibits strong generalization but struggles with complex articulations. 
Strong symmetry priors often cause hallucinated/missing limbs, overly straight 'canonicalized' shapes, and failures on extreme out-of-distribution poses (\eg the prone chimpanzee). 
In contrast, our predicted point clouds capture smooth deformations with natural curvature while disambiguating symmetric parts. 
Moreover, our template-fitting guarantees anatomical correctness and recovers challenging articulations.}
\label{fig:quali-eval-sam-trellis}
\end{figure}

\cref{fig:quali-eval-sam-trellis} demonstrates our scalability to other object categories, \ie chimpanzees and elephants, trained using real images. 
The image-to-3D foundation models struggle with complex articulations, frequently hallucinating limbs or producing overly straight, `canonicalized' shapes due to strong symmetry priors. 
In contrast, our template-fitting approach guarantees anatomical correctness, and performs better on challenging, asymmetric articulations.

\subsection{Ablations}
\label{sec:ablations}
\begin{figure}[h]
\centering
\includegraphics[width=\textwidth]{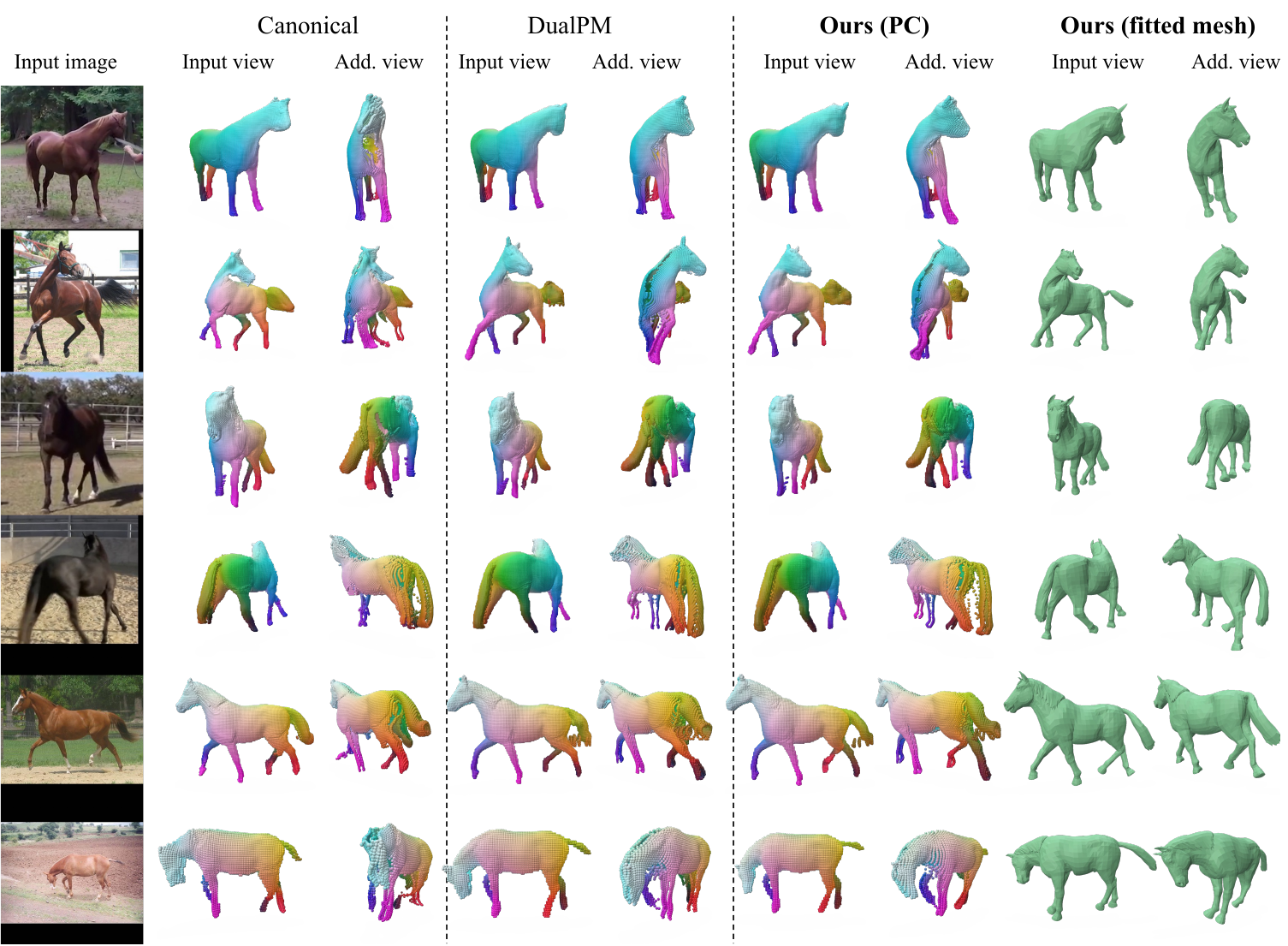}
\caption{Qualitative comparison on real-world horse images. We visualize point cloud predictions (input and novel views) and our extracted fitted meshes. While the `Canonical' baseline captures the coarse pose, it struggles with complex geometry. It exhibits noisy topology, missing parts, and mixed-up appendages (\eg confusing the rear legs). In contrast, our full model resolves these artifacts, yielding clean geometry that closely matches the fully supervised DualPM.}
\label{fig:ablation_2}
\end{figure}

In \cref{tab:ablations} we ablate the main components of our method on the \textit{Horse-Mixed} test split.
Our starting point is a \emph{Canonical (base)} model, trained exclusively on rest-pose renders with standard Gaussian-sampled camera viewpoints (see the supplementary material for details).
Replacing this with heavy-tailed Student's t camera sampling immediately yields substantial gains by exposing the initialization network to extreme, outlier viewpoints.
Next, we introduce our iterative pipeline (\emph{+ Refinement w/o reg.}), which extracts posed training data from images. 
While this provides a substantial performance leap, greedy point-cloud fitting produces structural artifacts. 
Adding geometric penalties (\emph{+ Regularization}) resolves this by enforcing physically plausible articulations.
Finally, \emph{+ Viewpoint augmentation} augments each fitted mesh with diverse novel viewpoints during rendering, forming our complete method.
\cref{fig:ablation_2} qualitatively compares the performance of our full model trained on real-world images to the canonical initialization and fully supervised DualPM.

We additionally report a \emph{Random poses baseline}, which trains the model on stochastically deformed meshes with unconstrained, unnatural articulations. Surprisingly, its relatively strong performance highlights that raw pose diversity alone provides a strong training signal (see supplementary material for details). Furthermore, we evaluate a \emph{Raw image baseline}, where we bypass our rendering step and train directly on the original image collection paired with the 3D supervision estimated from the previous iteration. As the initial mesh fitting is imperfect, this introduces natural inconsistencies between the 2D visual input and the 3D supervision. We observe that this setup fails to improve noticeably beyond its initialization, underlining the absolute necessity of strict 2D-3D consistency during training.

\begin{figure}[h]
\centering
\includegraphics[width=\textwidth]{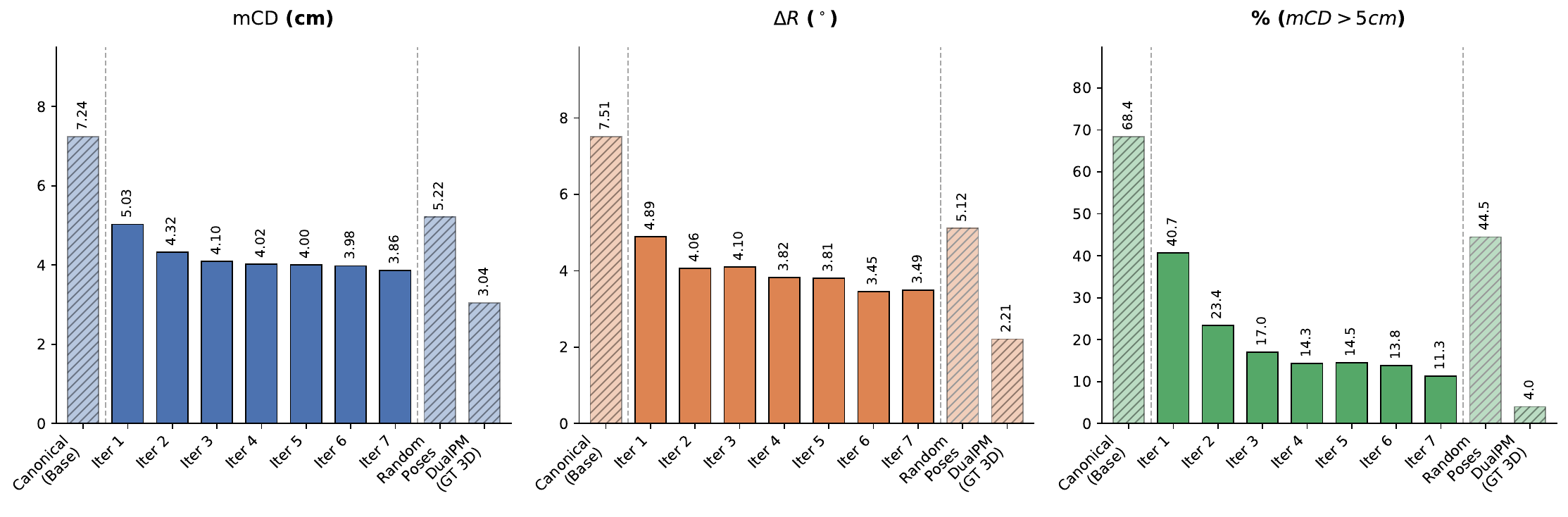}
\caption{Performance across refinement iterations. Errors drops significantly after the first iteration compared to the canonical baseline, then improves over successive iterations and plateaus while approaching the fully supervised DualPM (rightmost bars). Notably, the outlier metric \% (mCD$>$5cm) continues to show consistent improvements up through the seventh iteration.}
\label{fig:ablation-iter}
\end{figure}

\cref{fig:ablation-iter} presents an ablation on the number of iterations used in our approach. 
We observe a large jump in performance across all metrics after the first iteration of our method when compared to the canonical baseline.   
This improvement continues, approaching the performance of DualPM, but eventually plateaus. 

\begin{table}[tb]  
\centering
  \caption{Ablation study on the Horse-Mixed test split. Quantitative results evaluating the impact of removing individual pipeline components. 
  Lower scores are better. 
  }
  \vspace{-5pt}
\begin{tabular}{@{} l c c c @{}}
  \toprule
  Ablation variant & mCD (cm) & \% (mCD$>$5cm) & $\Delta R (\circ)$ \\
  \midrule
  DualPM (ground-truth 3D) & $3.04$ & $4.00$ & $2.21$ \\
  Random poses baseline & $5.22$ & $44.5$ & $5.12$ \\
  Raw image baseline & $6.96$ & $64.4$ & $8.38$ \\
\midrule
  Canonical (base) & $10.04$ & $92.0$ & $12.23$ \\
  + Student's $t$ camera sampling & $7.24$ & $68.4$ & $7.51$ \\
  + Refinement (w/o fitting reg.) & $5.59$ & $55.1$ & $6.73$ \\
  + Regularization in mesh fitting & $4.85$ & $35.0$ & $5.89$ \\
  + Viewpoint augmentation (`Ours') & $3.86$ & $11.3$ & $3.49$ \\
  \bottomrule
  \label{tab:ablations}
\end{tabular}
  \label{fig:ablation_main}
\end{table}

\subsection{Limitations}
\label{sec:limitations}
While significantly reducing the amount of 3D supervision needed compared to DualPM, our approach still needs some limited 3D data during training. 
Specifically, we require a rigged and skinned mesh in a canonical rest pose.  
However, in practice this is not difficult to acquire for common object categories and automated rigging methods could be used if only a mesh is available. 
As demonstrated by our experimental results, our approach successfully extracts pose diversity from real-world image collections to improve the predictor. However, \cref{tab:horse-robust} shows that a performance gap remains between training on real-world images, synthetic images, and ground-truth 3D. 
Finally, while our self-repulsion loss effectively prevents unnatural intersections, its current formulation activates broadly when body parts are in close proximity, which can occasionally hinder our ability to recover tight, compressed articulations where valid body parts rest against one another

\section{Conclusion}
\label{sec:conclusion}
Supervised learning-based approaches for 3D object reconstruction are only as effective as the data they are trained on. 
Unlike 2D images, which are available in abundance on the internet, diverse and high quality 3D datasets are much more challenging to create. 
To sidestep this issue, we introduced a new scalable and automated approach for 3D data generation called \methodname. 
Starting from only a single rigged mesh and 2D image collection as input, we showed that it is possible to bootstrap from this initial limited data, via an iterative generation approach, to create diverse and articulated 3D training data. 
We evaluated the effectiveness of our approach on the challenging task of articulated 3D object reconstruction from a single input image. 
We approach the performance of fully supervised methods across multiple categories, even when only provided with a fraction of the ground-truth 3D training data.

\clearpage

\vspace{10pt}
\noindent\textbf{Acknowledgements.} 
JZ was supported by the United Kingdom Research and Innovation (grant EP/S023208/1), UKRI Centre for Doctoral Training in Robotics and Autonomous Systems at the University of Edinburgh, School of Informatics.
OMA was supported by a Royal Society Research Grant. 
HB was supported by the EPSRC Visual AI grant EP/T028572/1.

\bibliographystyle{splncs04}

\clearpage
\appendix

\setcounter{table}{0}
\renewcommand{\thetable}{A\arabic{table}}
\setcounter{figure}{0}
\renewcommand{\thefigure}{A\arabic{figure}}

\title{Bootstrapping Articulated 3D Reconstruction from 2D Image Collections\\ -- Supplementary Material} 

\titlerunning{Bootstrapping Articulated 3D Reconstruction -- Supplementary Material}

\author{Jakub Zadrozny \and
Oisin Mac Aodha~\orcidlink{0000-0002-5787-5073} \and
Hakan Bilen~\orcidlink{0000-0002-6947-6918}}

\authorrunning{J.~Zadrozny et al.}

\institute{University of Edinburgh\\
\href{https://jakubzadrozny.github.io/bat3r}{\nolinkurl{jakubzadrozny.github.io/bat3r}}}

\maketitle

\input{supp_content}

\end{document}

%% file: supp_content.tex
\section{Additional results}

\subsection{Additional qualitative comparisons}
\begin{figure}[p]
\centering
\includegraphics[width=\textwidth]{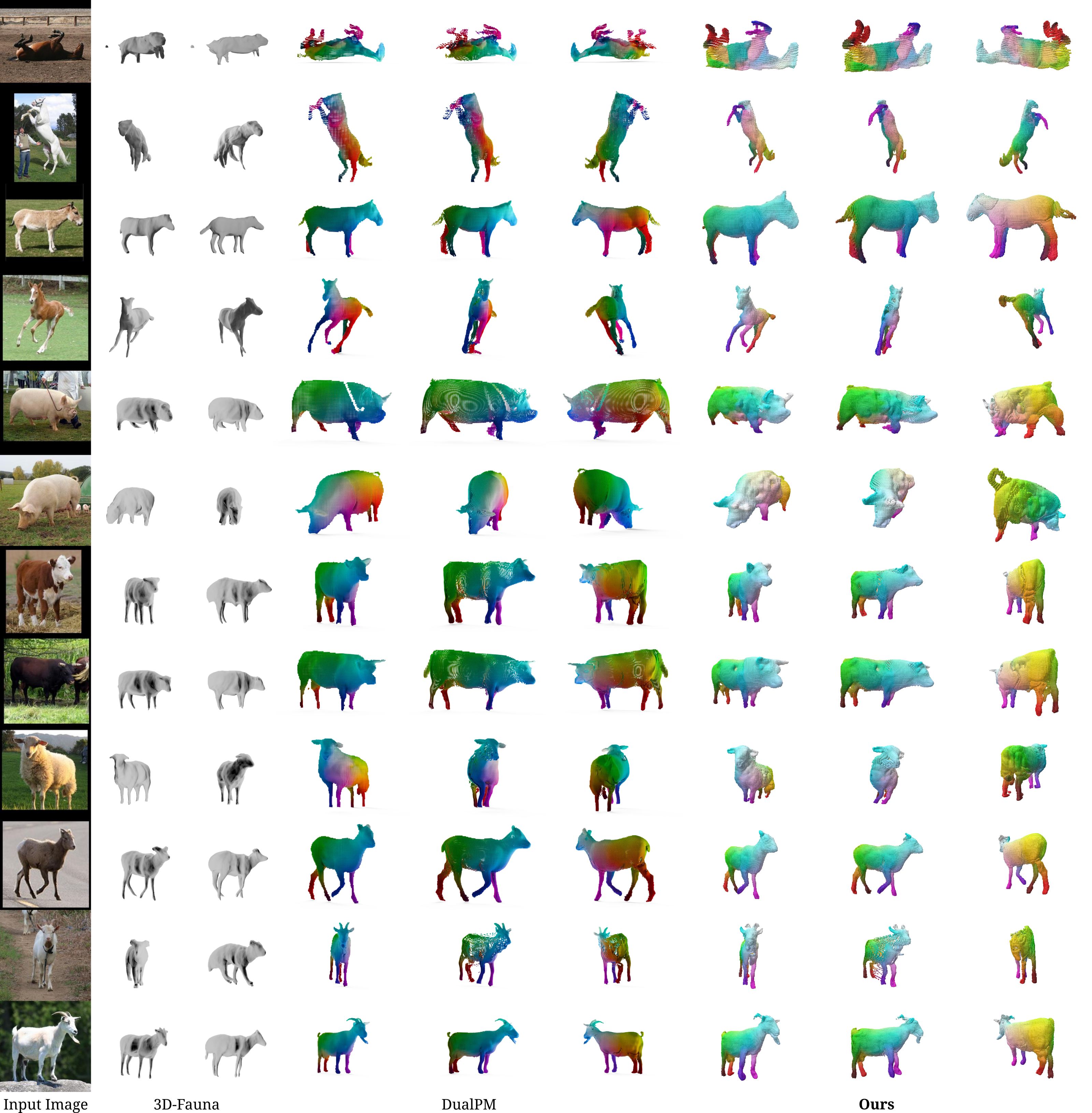}
\caption{
Qualitative comparisons to the unsupervised 3D-Fauna~\cite{li2024learning}, supervised DualPM~\cite{kaye2025dualpm}, and Ours. 
3D-Fauna struggles with challenging poses (see top two rows), whereas our approach obtains results comparable to the fully supervised DualPM. Our models in this comparison are trained on the synthetic images from DualPM following original splits.
}
\label{fig:result_1}
\end{figure}

See \cref{fig:result_1} for additional qualitative comparisons to the unsupervised 3D-Fauna~\cite{li2024learning} and fully-supervised DualPM~\cite{kaye2025dualpm} on the horse, cow, and sheep categories. This comparison uses the same models as \cref{tab:main_res}, \ie trained on the synthetic images from DualPM following original splits.

\begin{figure}[t]
\centering
\includegraphics[width=\textwidth]{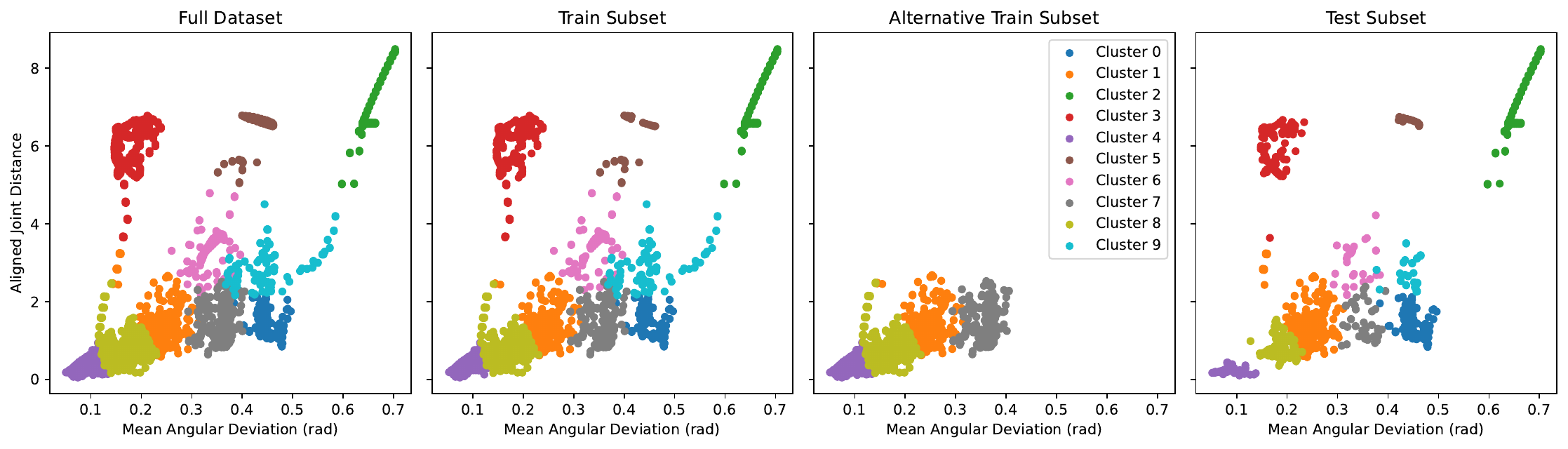}
\caption{\textbf{Pose diversity across dataset splits.} 2D scatter plots mapping dataset samples by their aligned joint distance ($f_{dist}$) and mean angular deviation ($f_{ang}$). We visualize the geometric distribution of poses across the Full Dataset, the stratified Base Train Subset, the restricted Alternative Train Set, and the held-out Test Set. Crucially, the Alternative Train Set deliberately excludes highly deformed and out-of-distribution samples (clusters 0, 2, 3, 5, 6, and 9), isolating the core articulations to evaluate model generalization.}
\label{fig:clusters-scatter}
\end{figure}

\subsection{Performance across different object pose clusters}
\label{sec:clustering}
To get a deeper understanding of the influence of object pose diversity on the performance of the trained 2D-to-3D predictor we partition the diverse range of poses in the dataset of~\cite{kaye2025dualpm} into ten distinct, semantic clusters using the K-Means algorithm. To ensure the clustering is invariant to the global camera viewpoint and focuses strictly on the intrinsic body configuration, we extract three key geometric features for each pose.

Let $J$ denote the set of skeletal joints. For a given pose defined by the 
local joint transforms $T_j$ we first compute the excess local joint rotations (from the rest pose) $\theta_j$. Let $\theta_k$ represent the axis-angle rotation vector of the $j$-th joint, let $\theta$ be the stacked local joint rotations $\theta_j$, and let $x_j(\theta)$ and $x^c_j$ denote the $j$-th joint 3D position in the articulated state and the canonical rest pose, respectively. For each sample in the dataset we compute the following feature vector $f = [f_{ang}, f_{sq}, f_{dist}]^\top$, where we define:
\begin{itemize}
    \item \textbf{Mean angular deviation:} $f_{ang} = \frac{1}{|J|} \sum_{k \in J} \|\theta_k\|_2$. This captures the average magnitude of local excess joint rotations across the entire skeleton.
    \item \textbf{Mean squared angular deviation:} $f_{sq} = \frac{1}{|J|} \sum_{k \in J} \|\theta_k\|_2^2$. By penalizing larger rotations more heavily, this feature effectively between poses characterized by widespread, gradual bending versus those exhibiting localized, extreme articulations.
    \item \textbf{Aligned joint distance:} $f_{dist} = \sum_{k \in J} \| T_\textrm{align}(x_k(\theta)) - x^c_k \|_2$. We apply the Kabsch algorithm to find the optimal rigid transformation $T_\textrm{align}$ (comprising only rotation and translation) that aligns the articulated joints $x_k(\theta)$ to their canonical counterparts $x^c_k$. This feature measures the residual Euclidean distance after alignment, capturing the overall spatial deformation of the skeleton.
\end{itemize}

We then standardize the features to zero mean and unit variance, and apply K-Means clustering ($K=10$) to assign each pose to its respective articulation cluster. \cref{fig:clusters-scatter} 
contains a scatter plots mapping dataset samples by their aligned joint distance ($f_{dist}$) and mean angular deviation ($f_{ang}$).

\subsubsection{\bf Dataset splitting and generalization sets.}
Here we measure DualPM's capacity to generalize to unseen or underrepresented articulations by constructing two distinct training set variants from the full dataset.
First, we define a \textit{Base Train Set}, which is generated by sampling from the entire dataset, stratified by the geometric K-Means cluster IDs (as defined in \cref{sec:clustering}). This ensures a proportional representation of all discovered articulation modes. 
Second, we construct an \textit{Alternative Train Set} which represents the designed to test generalization on non-trivial, intricate, out-of-distribution poses. For this set, we explicitly exclude all samples belonging to clusters which represent poses significantly differing from the rest pose (cluster IDs \{0, 2, 3, 5, 6, 9\}). Since these clusters are underrepresented in the general dataset, they only comprise 5,000 training samples, \ie $\approx31.5\%$ of the original training set. 
We retain all 10,872 samples from the allowed clusters (\{1, 4, 7, 8\}) that were already present in the base train set and compensate for the total volume of the alternative train set by sampling additional instances from the allowed clusters to match the size of the original train set (15,872 samples). This yields a deliberate overlap of exactly 10,872 samples (approximately $68.5\%$) between the two training splits. The exact composition of our overall dataset, along with the two training splits and the held-out test set (non-intersecting with neither the original train set nor the alternative), is detailed in \cref{tab:cluster_dist}.

\subsubsection{Results.}
\begin{figure}
\centering
\includegraphics[width=0.95\textwidth]{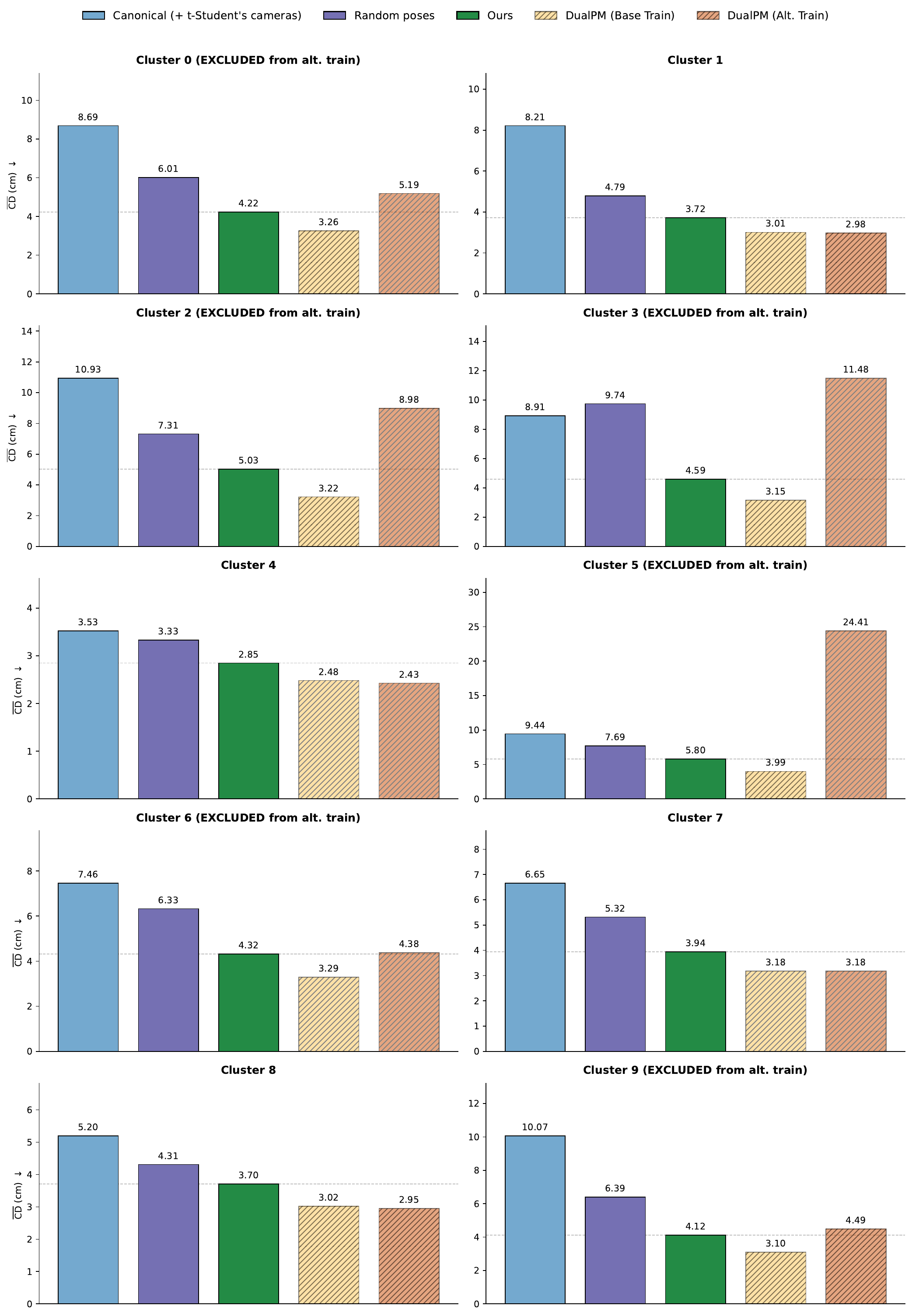}
\caption{mCD (cm) breakdown by cluster ID. While fully-supervised models like DualPM struggle to generalize to out-of-distribution poses (for the Alternative Train Set), our scalable refinement strategy (`Ours') leverages diverse unannotated 2D images to maintain high accuracy across all clusters.}
\label{fig:clusters-barplots}
\end{figure}

\cref{fig:clusters-barplots} reports the per-cluster performance of our method compared to several baselines: our canonical-only initialization, the random poses baseline, and the fully-supervised DualPM trained on both the Base (wide) and Alternative (narrow) splits.

First, we note that generalization of DualPM to in-distribution poses (\eg clusters 0, 6, and 9 are closely related to clusters 1, 7, 8) is good. Nevertheless, providing our approach with the wider variety of training images (which is easier to collect at scale compared to ground-truth poses) compensates the lack of ground-truth poses in these cases (our approach slightly outperforms the gold DualPM trained on the alternative test set).

We observe that DualPM generalizes reasonably well to structurally similar in-distribution poses (\eg the held-out clusters 0, 6, and 9 share geometric similarities with the permitted training clusters 1, 7, and 8). Nevertheless, our proposed refinement, which leverages a wider variety of unannotated 2D training images, effectively compensates for the lack of ground-truth 3D supervision. On these structurally related clusters, our approach slightly outperforms the DualPM model trained on the restricted Alternative set.

Crucially, this experiment highlights that DualPM struggles to generalize to strictly out-of-distribution (OOD) articulations (\eg clusters 2, 3, and 5, though we note cluster 5 also exhibits a significant viewpoint distribution shift). On these OOD clusters, our approach significantly outperforms the DualPM baseline trained on the Alternative train set. Interestingly, for clusters 2 and 5, even our naive canonical-trained model surpasses the narrow DualPM variant, indicating that explicit 3D supervision on limited articulation modes makes the network susceptible to pose-related overfitting.

Consequently, adapting fully-supervised methods like DualPM to novel categories necessitates a comprehensive dataset of manually authored 3D articulations, which is a highly expensive and time-consuming process. Omissions in the training distribution directly bottlenecks downstream performance. In contrast, our refinement strategy provides a scalable alternative: by deriving geometric priors directly from widely available, diverse 2D image collections, we yield a more robust and generalized 3D estimator.

\begin{figure}
\centering
\includegraphics[width=\textwidth]{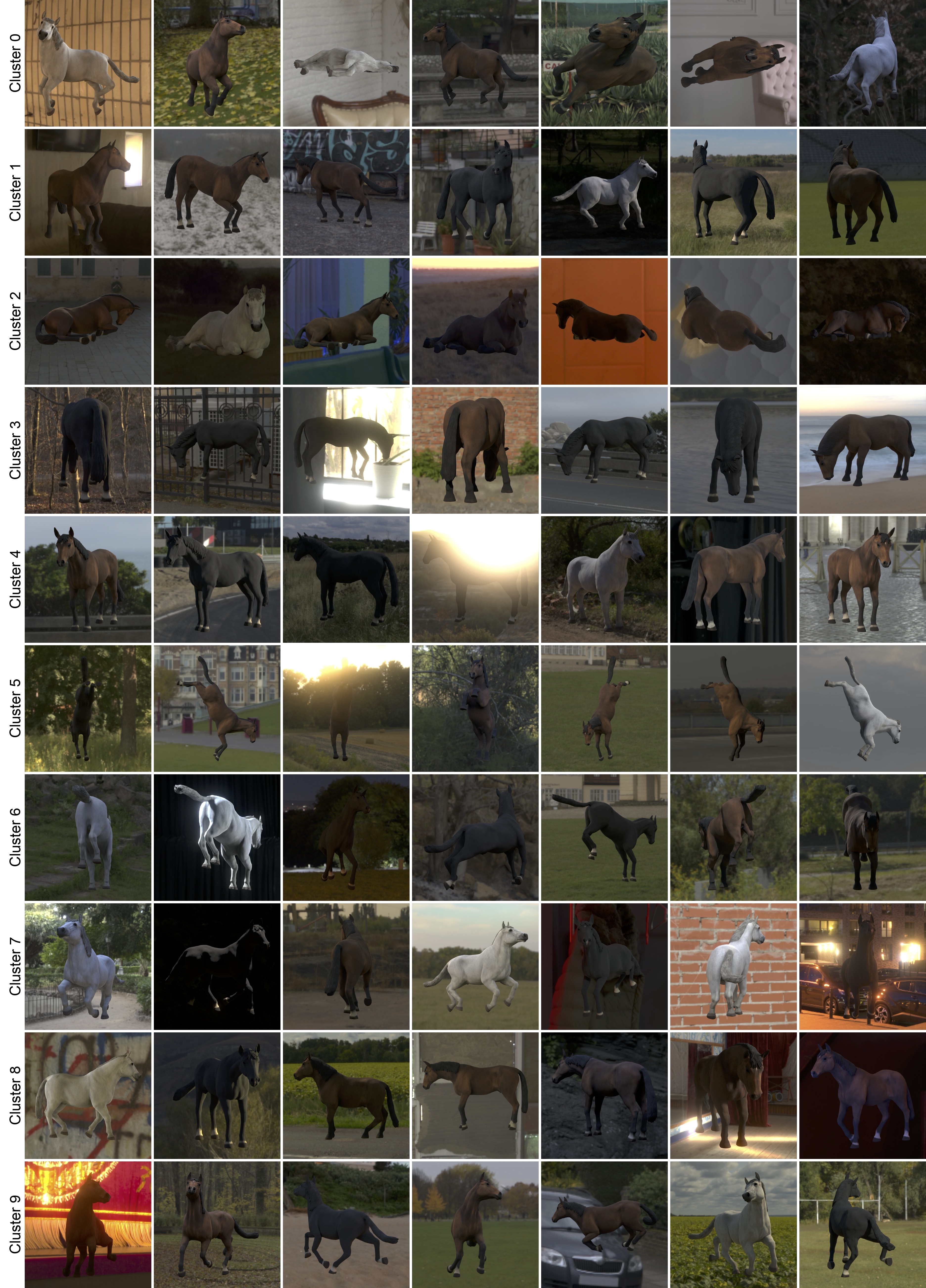}
\caption{Examples of synthetic images from the DualPM~\cite{kaye2025dualpm} dataset split by the assigned cluster.}
\label{fig:clusters-grid}
\end{figure}

Furthermore, the per-cluster breakdown confirms the intuitive premise that the advantages of our refinement process (as well as explicit 3D ground truth) scale with increasing articulation complexity. As depicted in \cref{fig:clusters-barplots}, our approach achieves its most substantial margins of improvement over the canonical and random-pose baselines on clusters 2, 3, and 9. Visualized in \cref{fig:clusters-grid}, these clusters are characterized by highly non-trivial, extreme articulations. Conversely, for clusters 4 and 8, which predominantly feature standing poses closely resembling the canonical rest state, all evaluated variants achieve comparable, accurate alignments.

\begin{table}[tb]  
\centering
\caption{Distribution of articulation modes across dataset splits. For each cluster, we report the raw sample count alongside its relative percentage within that specific split. The Alternative Train split entirely excludes six clusters while upsampling the remaining four to maintain a consistent total training size.}
\resizebox{0.8\textwidth}{!}{
\begin{tabular}{@{} c | r r r r @{}}
  \toprule
  \textbf{Cluster ID} & \textbf{Overall} & \textbf{Base Train} & \textbf{Alt.~Train} & \textbf{Test} \\
  \midrule
  0 & $9,129$ ($9.1\%$) & $1,548$ ($9.8\%$) & $0$ ($0.0\%$) & $1,895$ ($9.8\%$) \\
  1 & $26,427$ ($26.4\%$) & $4,390$ ($27.7\%$) & $6,276$ ($39.5\%$) & $5,386$ ($27.7\%$) \\
  2 & $3,124$ ($3.1\%$) & $613$ ($3.9\%$) & $0$ ($0.0\%$) & $754$ ($3.9\%$) \\
  3 & $3,184$ ($3.2\%$) & $522$ ($3.3\%$) & $0$ ($0.0\%$) & $643$ ($3.3\%$) \\
  4 & $13,335$ ($13.3\%$) & $2,008$ ($12.7\%$) & $3,012$ ($19.0\%$) & $2,470$ ($12.7\%$) \\
  5 & $2,560$ ($2.6\%$) & $256$ ($1.6\%$) & $0$ ($0.0\%$) & $305$ ($1.6\%$) \\
  6 & $5,595$ ($5.6\%$) & $870$ ($5.5\%$) & $0$ ($0.0\%$) & $1,057$ ($5.4\%$) \\
  7 & $12,587$ ($12.6\%$) & $1,946$ ($12.3\%$) & $2,881$ ($18.2\%$) & $2,375$ ($12.2\%$) \\
  8 & $16,008$ ($16.0\%$) & $2,528$ ($15.9\%$) & $3,703$ ($23.3\%$) & $3,097$ ($15.9\%$) \\
  9 & $8,051$ ($8.1\%$) & $1,191$ ($7.5\%$) & $0$ ($0.0\%$) & $1,450$ ($7.5\%$) \\
  \midrule
  \textbf{Total} & \textbf{$100,000$} ($100\%$) & \textbf{$15,872$} ($100\%$) & \textbf{$15,872$} ($100\%$) & \textbf{$19,432$} ($100\%$) \\
  \bottomrule
\end{tabular}
}
\label{tab:cluster_dist}
\end{table}

\section{Ablations}

\subsection{Further details on ablation variants}
\begin{figure}[h]
\centering
\includegraphics[width=\textwidth]{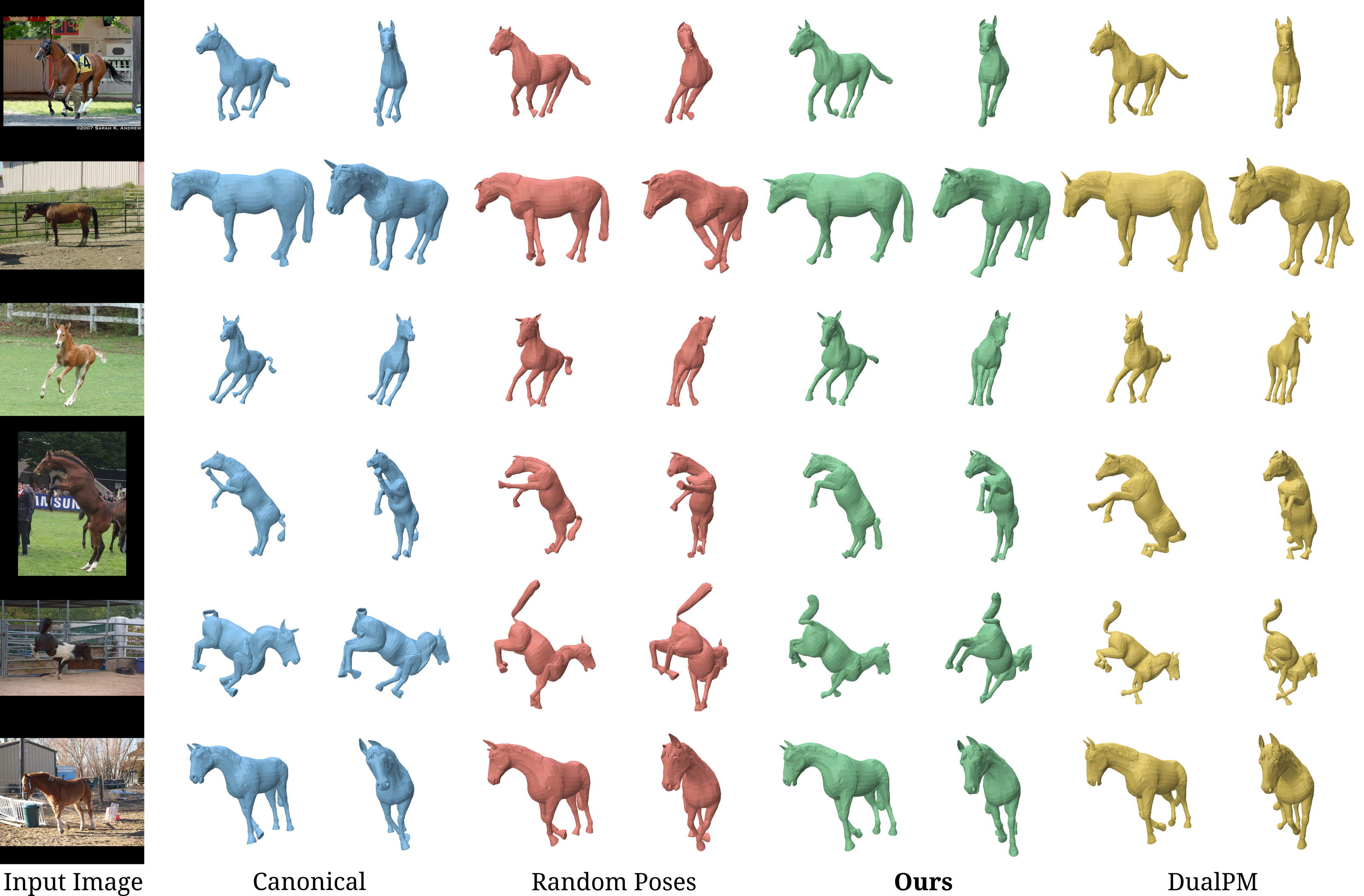}
\caption{
Qualitative comparisons to different variants of our model. 
The canonical model (blue), despite being only trained on data from a single mesh, is surprisingly effective at generalizing to unusual poses (see fourth row). 
This is likely thanks to the strong prior provided by the DINOv2~\cite{oquab2023dinov2} features used by DualPM. 
Similarly, the model trained on random poses (red) performs well, but upon close inspection there are noticeable artifacts, \eg the crossed legs in the second row. 
Our results (green) are comparable to DualPM's (yellow), but trained on a fraction of the 3D supervision. 
}
\label{fig:ablation_1}
\end{figure}

In this section, we detail the specific ablation variants evaluated in \cref{fig:ablation_main}. Moreover, \cref{fig:ablation_1} shows a qualitative comparison of different ablation variants of our approach.

Our baseline, \emph{`Canonical (base)'}, model consists of a 3D prediction module trained exclusively on synthetic renders of the provided mesh in its canonical (\ie rest) pose. The camera poses for this base variant (defined in spherical coordinates) are determined as follows: the azimuth angle is sampled uniformly across the full $360^\circ$ range, while the elevation and roll angles are drawn from a standard Gaussian distribution fitted to the empirical poses of the DualPM dataset, as detailed in \cref{sec:viewpoint-analysis}. The camera is centered at the mesh, with the radius dynamically adjusted to ensure the subject fills the image frame, leaving a relative bounding margin uniformly sampled from $[0.1, 0.25]$.

We then evaluate an updated canonical baseline \emph{(`+ Student's $t$ camera sampling')} that replaces the Gaussian sampling with the heavy-tailed Student's $t$-distribution (see \cref{sec:viewpoint-analysis}). As shown in the results, this adjustment alone yields a substantial performance improvement by better exposing the network to extreme, outlier viewpoints during initialization.

Next, we introduce our iterative refinement pipeline \emph{(`+ Refinement (w/o fitting reg.)')} as described in \cref{sec:method}. In this ablation, the target mesh fitting is performed in a purely greedy manner, \ie omitting all geometric regularization loss terms. Furthermore, we do not utilize the viewpoint augmentation step (\cref{sec:camera_mult}). Instead, each training image generates exactly one paired 2D-3D training instance using the camera pose estimated directly from the input image. 

Building on this, the \emph{`+ Regularization in mesh fitting'} variant reintroduces the structural regularization losses during the fitting phase to encourage physically plausible mesh states. 

Finally, the \emph{`+ Viewpoint augmentation'} variant augments each fitted mesh with diverse novel viewpoints during rendering which we describe below to address front/rear viewpoint performance. This final configuration represents our complete, proposed methodology (\ie `Ours').

\subsection{Random poses baseline}
\label{sec:random-poses}
\Cref{fig:ablation_main} in the main text compares our proposed approach against a naive baseline (`Random poses baseline`), which trains the DualPM architecture entirely on stochastically generated mesh articulations. For this baseline, we synthesize a training dataset of $N=45,000$ samples. As detailed in \cref{sec:method}, each pose is parameterized by the excess joint rotations relative to the rest pose, parameterized as axis-angle vectors.

To sample a single random pose, we compute the rotation for each joint independently. First, a rotation axis is sampled uniformly at random from the unit sphere. The magnitude of this rotation (the angle) is then drawn from a Half-Normal distribution truncated at $\pi$. The variance of this distribution is governed by a shared pose-level complexity parameter $\sigma$, defined as $\sigma = \sigma_1 + t(\sigma_2 - \sigma_1)$. To ensure the generated dataset contains a natural, continuous blend of subtle, near-rest poses and highly deformed articulations, the interpolation factor $t \in [0, 1]$ is sampled from a $\text{Beta}(1, 3)$ distribution for each pose, with empirical bounds set to $\sigma_1 = 0.1$ and $\sigma_2 = 0.5$. To filter out highly degenerate or particularly extreme outliers, we evaluate the mean angular deviation of each generated pose (as defined in \cref{sec:clustering}) and discard all samples falling within the upper quartile ($25\%$). Finally, in this `Random poses baseline` experiment we use the same camera pose sampling strategy as in the `+ Student's $t$ camera sampling' ablation.

\begin{figure}[t]
\centering
\includegraphics[width=\textwidth]{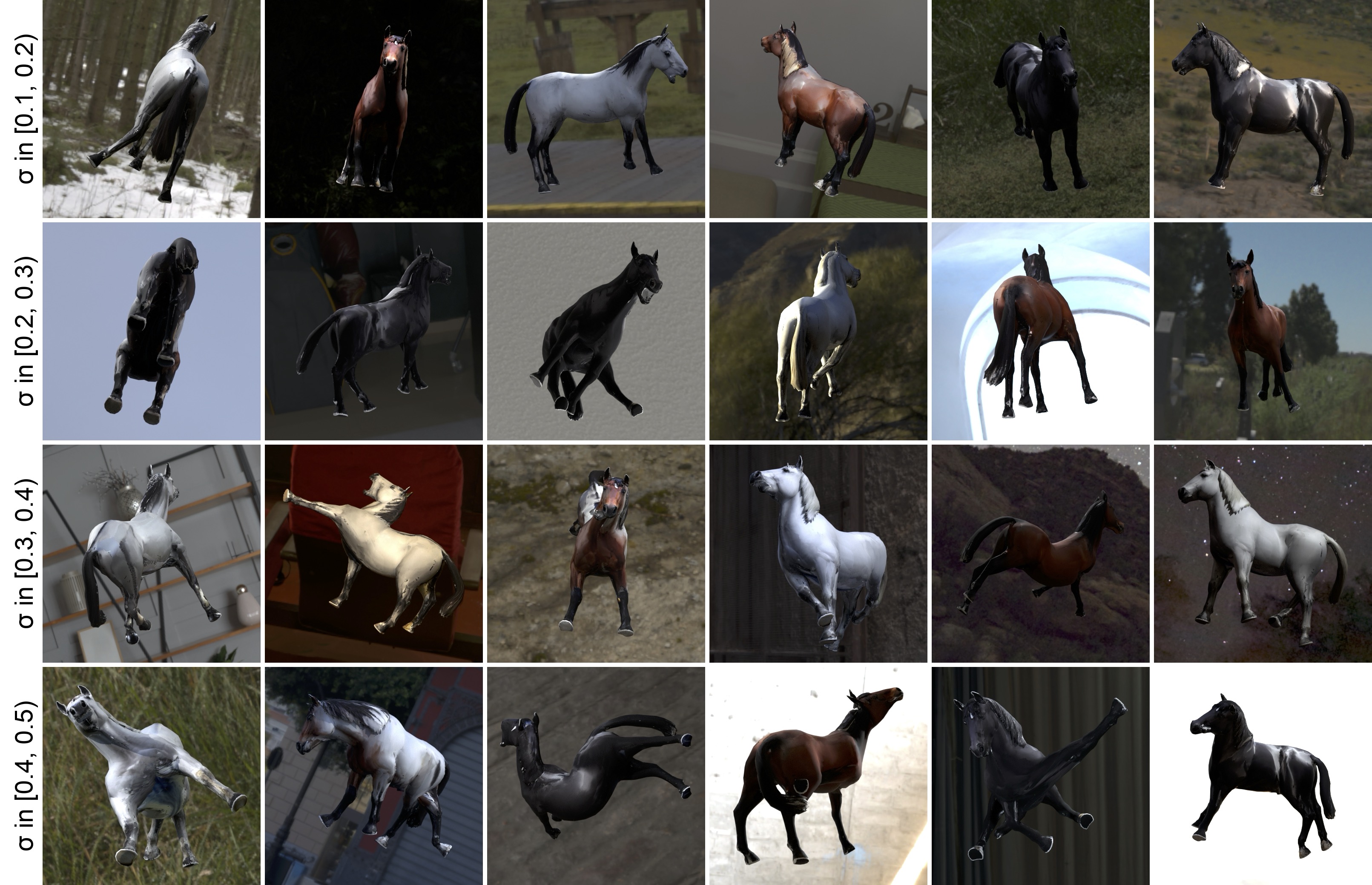}
\caption{
Examples of training samples generated based on random articulations (see \cref{sec:random-poses}). The $\sigma$ of the Half-Normal distribution used to sample joint angles is shown on the left.}
\label{fig:rand-poses-grid}
\end{figure}

It is important to note that these stochastically sampled poses are entirely unconstrained by biomechanical limits. Consequently, the resulting articulated meshes frequently exhibit severe physical artifacts, including self-intersections, unnatural stretching, and volume loss (see visual examples in \cref{fig:rand-poses-grid}). Despite these severe geometric distortions, this baseline achieves unexpectedly competitive quantitative performance (see \cref{fig:ablation_main}). See \cref{fig:ablation_1} for a qualitative comparison to other ablation variants.
This highlights the inherent robustness of the underlying architecture when exposed to high-variance spatial data, further validating the necessity of broad pose diversity during training.

\subsection{Camera viewpoint analysis for robust synthetic sampling}
\label{sec:viewpoint-analysis}
To systematically analyze the distribution of camera poses, we map each world-space camera into a normalized, body-centric spherical coordinate system invariant to the global orientation of the articulated mesh. 

First, we establish a local orthogonal coordinate frame anchored to the subject's skeletal joints. This requires the manual definition of an origin joint, alongside two pairs of joints to dictate the forward axis and the horizontal reference plane. For example, on the horse mesh, a natural configuration utilizes the middle of the spine as the origin, two adjacent spine joints for the forward axis, and the left and right hips to define the horizontal plane. These initial directional vectors serve as a rough basis for subsequent orthogonalization, meaning the selected joint pairs need not be strictly coplanar or intersecting.

\begin{figure}[h]
\centering
\includegraphics[width=\textwidth]{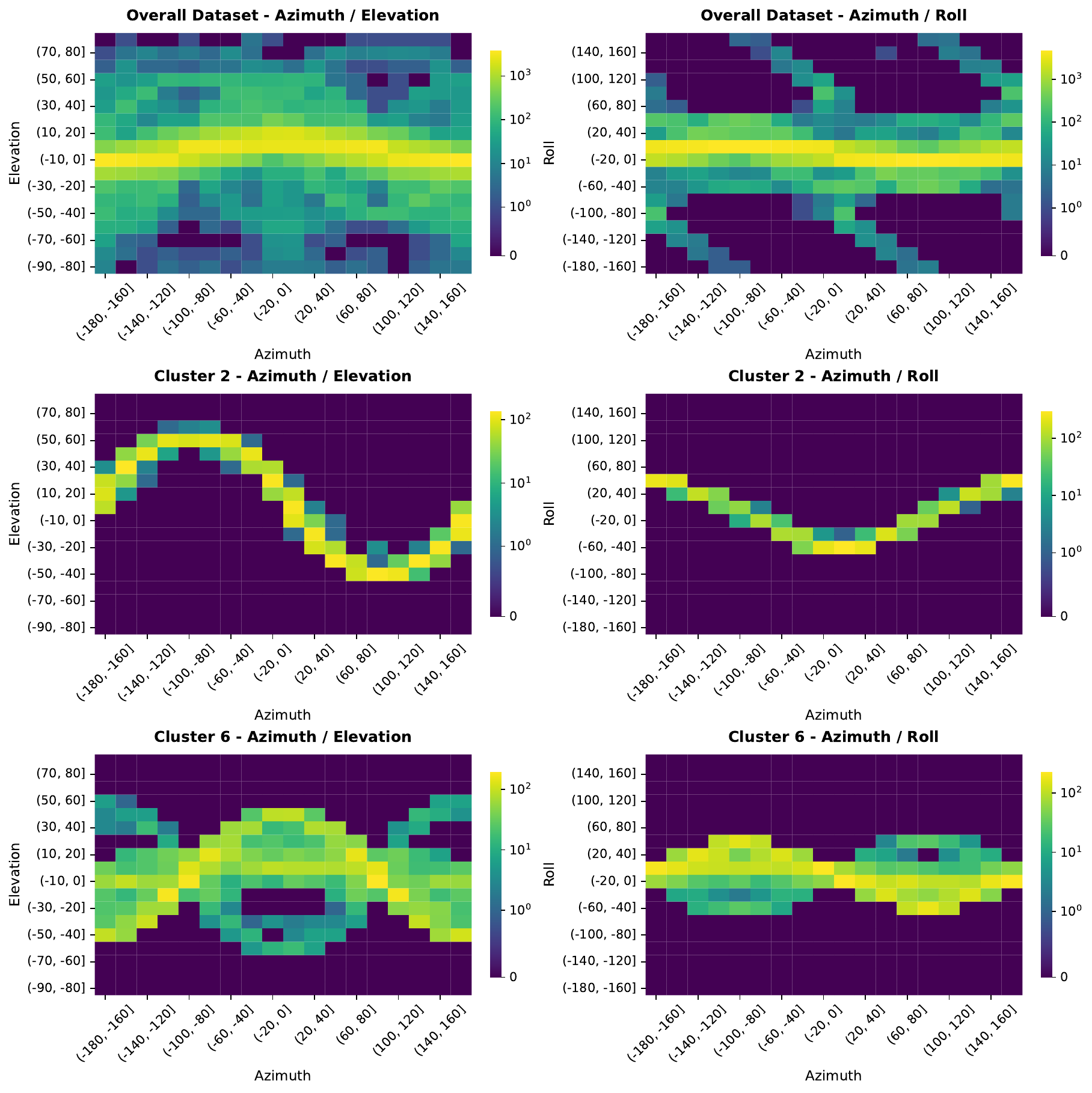}
\caption{
Camera viewpoint distribution heatmaps (in log-scale) in the DualPM~\cite{kaye2025dualpm} dataset for the entire dataset and selected clusters.}
\label{fig:camera-2d-hist}
\end{figure}

Let $\Delta P$ denote the Cartesian position of the camera relative to the origin joint (both in world space). We project $\Delta P$ into the local body frame to extract the standard spherical camera parameters: azimuth, elevation, roll, and radius. The empirical distribution of camera poses within the synthetically generated DualPM dataset is visualized in \cref{fig:camera-2d-hist}.  
As the camera radius is consistently scaled to ensure the subject fills the image frame (as shown in \cref{fig:clusters-grid}), we restrict our quantitative analysis to the angular parameters: azimuth, elevation, and roll. In this convention, an azimuth of $0^\circ$ designates a direct frontal view, $\pm 90^\circ$ correspond to the lateral profiles, and $\pm 180^\circ$ indicates a strict rear view. While the azimuth angle exhibits a generally uniform distribution, its joint distribution with elevation and roll reveals notable structural patterns. 

A primary observation from \cref{fig:camera-2d-hist} is the substantial presence of unusual viewpoints within the dataset. Furthermore, a per-cluster breakdown (\eg clusters 2 and 6 in \cref{fig:camera-2d-hist}) indicates the presence of explicitly hand-crafted camera trajectories that deviate significantly from a neutral, zero-elevation, and zero-roll perspective.

\begin{figure}[h]
\centering
\includegraphics[width=\textwidth]{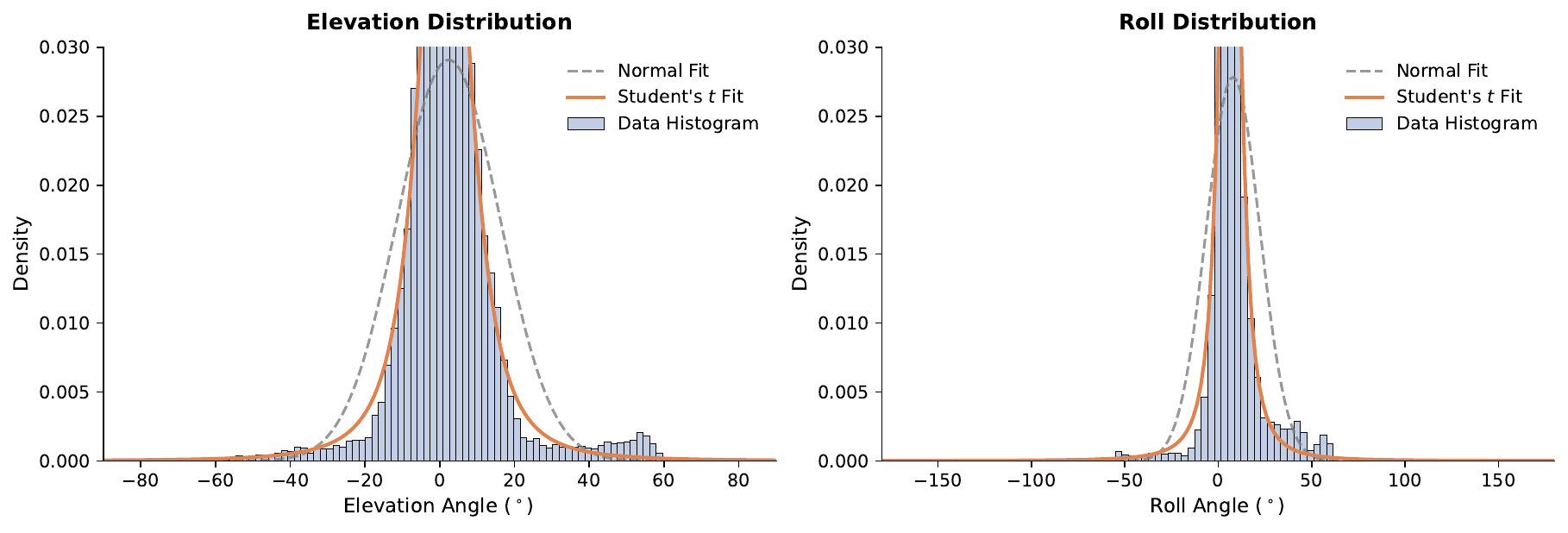}
\caption{\textbf{Parametric fitting of camera pose distributions.} Density histograms of the empirical camera elevation (left) and roll (right) angles from the DualPM dataset, overlaid with Normal (Gaussian) and Student's $t$-distribution fits. The Student's $t$-distribution captures the heavy tails of the empirical data better compared to the standard Gaussian, motivating its use to robustly sample extreme viewpoints during our synthetic dataset initialization.}
\label{fig:camera-parametric-fit}
\end{figure}

Recall that to initialize our iterative refinement pipeline, we synthesize a dataset by rendering the canonical mesh from diverse viewpoints. To ensure an adequate representation of outlier views while preventing strict overfitting to the exact DualPM camera views, we sample the azimuth angle uniformly, whereas the elevation and roll angles are drawn from a Student's $t$-distribution parameterized by fitting the empirical DualPM data. As illustrated in \cref{fig:camera-parametric-fit}, the Student's $t$-distribution effectively captures the heavy-tailed nature of the empirical viewpoint data, clearly outperforming a standard Gaussian fit. The quantitative impact of this heavy-tailed sampling strategy is validated in our ablations (see \cref{fig:ablation_main}), demonstrating that robustly representing extreme viewpoints during canonical initialization is critical for the stability and accuracy of the final 3D estimator.

\begin{figure}[h]
\centering
\includegraphics[width=\textwidth]{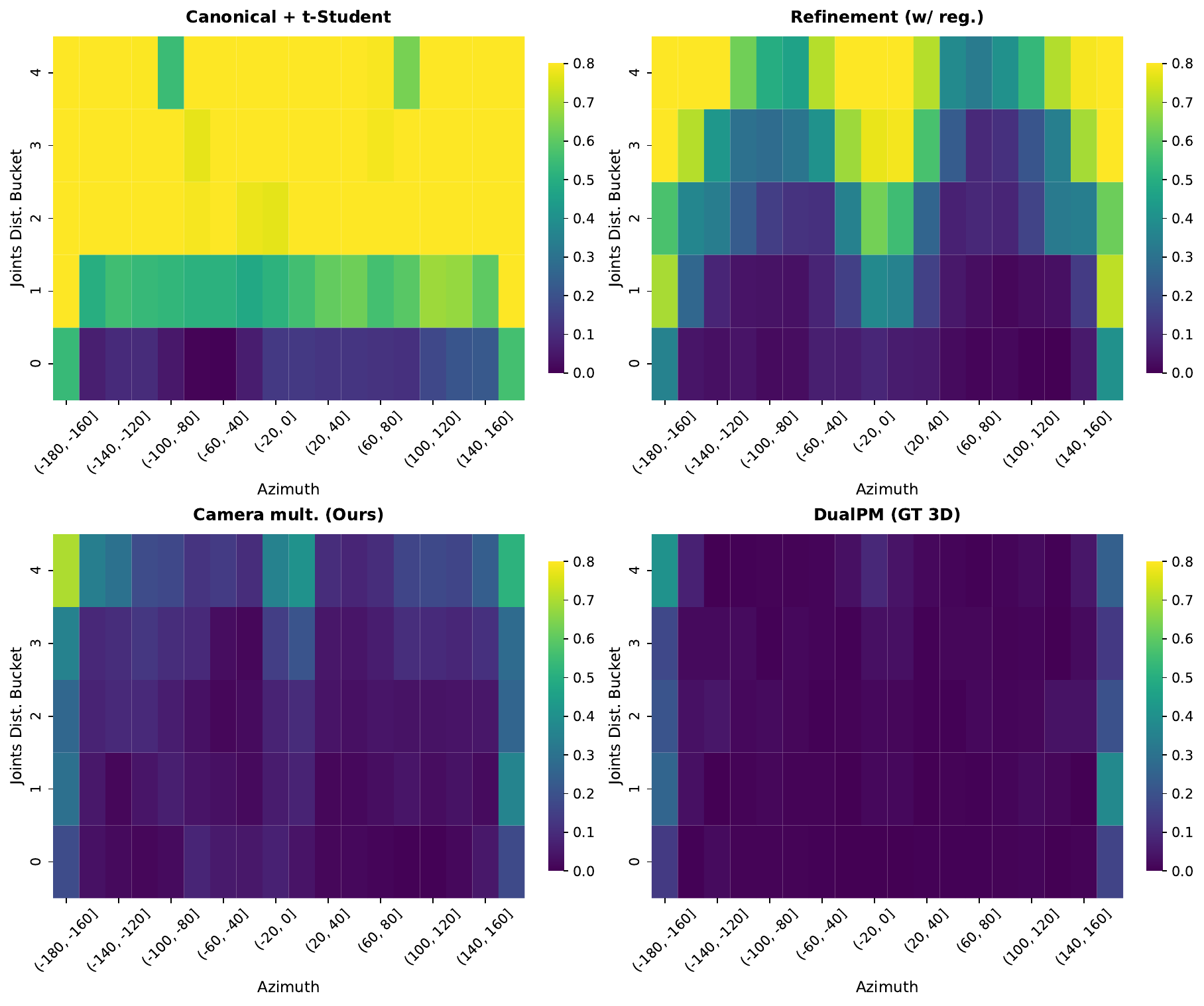}
\caption{\textbf{Reconstruction failure rates across viewpoints and pose complexity.} Heatmaps illustrating the fraction of test samples exceeding a 5cm error threshold (\% (mCD$>$5cm)) across different model variants. The horizontal axis represents the camera azimuth angle ($0^\circ$ denotes frontal, $\pm 90^\circ$ lateral), while the vertical axis denotes pose complexity (binned aligned joint distance, increasing with Y coordinate). While the base refinement strategy significantly reduces errors for lateral views, frontal and rear perspectives remain challenging. Our proposed viewpoint augmentation strategy effectively bridges this gap by leveraging reliable lateral predictions to bootstrap the underperforming angles.}
\label{fig:error_rate_heatmap_grid}
\end{figure}

\subsection{Viewpoint augmentation}
\label{sec:camera_mult}

\cref{fig:error_rate_heatmap_grid} shows the fraction of test samples exceeding a mean 5cm error threshold (effectively representing significant reconstruction failures), stratified by the ground-truth camera azimuth angle and the binned aligned joint distance (which serves as a proxy for pose complexity, as detailed in \cref{sec:clustering}). A key observation is that our base refinement process (prior to viewpoint augmentation) yields particularly effective improvements for lateral views (corresponding to an azimuth of $\pm 90^\circ$) relative to the canonical-only baseline, outpacing the gains observed for frontal or rear perspectives.

A natural approach to mitigate this performance discrepancy is to leverage the highly accurate lateral predictions to bootstrap more reliable frontal and rear training samples. Specifically, we dynamically sample supplementary views based on the primary camera perspective during training. For a given base viewpoint categorized broadly as frontal, rear, or lateral we sample $N \in \{1, 2\}$ additional viewpoints. Crucially, this strategy relies entirely on our predicted camera poses rather than ground-truth annotations. 
As shown in \cref{fig:ablation_main}, our estimated poses exhibit an average rotation error of only $3.5^\circ$, making them highly reliable for this routing. To mildly upsample historically challenging angles, we assign a higher probability ($65\%$) of generating two supplementary views when the base view is lateral, compared to a lower probability ($35\%$) when starting from a frontal or rear view.

The specific categories of these supplementary viewpoints are drawn from a conditional probability distribution designed to favor complementary angles. Namely, we heavily penalize sampling the identical view type again and introduce a soft bias that encourages lateral base views to sample frontal or rear perspectives. Importantly, this sampling bias is applied gently. While we explicitly favor synthesizing frontal and rear views from lateral inputs, we strictly maintain a non-zero probability for all cross-viewpoint combinations (\eg generating lateral views from frontal ones, or frontal from frontal). This soft conditioning ensures that we successfully target the challenging angles without severely distorting the natural viewpoint distribution of the overall dataset.

\section{Additional implementation details}
\label{sec:implementation_details}

\noindent{\bf Mesh fitting and regularization.} 
During the mesh fitting stage, we employ the Adam optimizer for 200 total steps. The initial learning rate is set to $2 \times 10^{-2}$ for the joint rotation and bone scale (if they are enabled) parameters and $1 \times 10^{-3}$ for the rigid transformation parameters. To ensure stable convergence, the learning rate for the joint rotations is halved after every 100 steps. 

For synthetic images, the empirical weights for the regularization terms defined in Eq.~(3) are set as follows: the angle loss weight $\lambda_{angle} = 1 \times 10^{-5}$, the repulsion weight $\lambda_{rep} = 1.0$, the edge length loss weight $\lambda_{edge} = 100$, and the local volume loss weight $\lambda_{vol} = 20$. Bone scaling is disabled entirely for synthetic images, so $\lambda_{scale} = 0$. Because real-world images contain more instance variability and are more noisy, their mesh fitting setup is slightly different. First, we run the optimizer for 300 total steps. Then, the real-world parameters (shared by horse, chimpanzee, elephant categories) are $\lambda_{angle} = 3 \times 10^{-5}$, $\lambda_{rep} = 5.0$, $\lambda_{edge} = 3000$, and the local volume loss stays at $\lambda_{vol} = 20$. We also use a non-zero $\lambda_{scale} = 0.01$ for real-world images. Note the loss terms operate in different range and these weights generally compensate for that.

To avoid test-set leakage, these synthetic hyperparameters were determined via visual inspection of the fit quality and physical realism on a small subset of the \textit{horse} training data. These values demonstrated strong generalizability and were consequently directly carried over to the \textit{cow} and \textit{sheep} categories without further tuning. The real-world mesh fitting parameters were fine-tuned from the synthetic ones by visual inspection of a couple dozen samples from the three categories.

\noindent{\bf Point map predictor training.}
To stabilize training, we accumulate gradients over two batches, slightly increase the Adam~\cite{kingma2014adam} momentum, apply a weight decay of $0.01$, and utilize a sequential learning rate scheduler comprising a linear warmup (1K steps) followed by cosine annealing. While we maintain the original base learning rate, we reduce the training duration to 60K steps per refinement iteration, which helps offset the computational overhead of gradient accumulation. Retraining the baseline DualPM under this stabilized regime yielded only marginal improvements, confirming these modifications are specifically necessary for handling noisy, self-supervised targets. 

\noindent{\bf Textures.}
To synthesize paired 2D-3D data during the iterative refinement procedure, the meshes must be textured. We use 1-2 manually designed textures bundled with the canonical meshes for cows, sheep, chimps, and elephants, and employ 4 textures automatically generated using Text2Tex~\cite{chen2023text2tex}.
Our approach inherits the feature extractor based on~\cite{zhang2023tale} from DualPM, which ensures strong generalization to appearance.
We set the background texture and illumination sampling using the same set of HDRI environment maps as DualPM.

\noindent{\bf Synthetic rendering pipeline.} 
We utilize Blender as our primary rendering engine. Environmental lighting and background the sampled HDRI environment maps (as in~\cite{kaye2025dualpm}). To texture the articulated meshes, we apply four texture maps generated using Text2Tex~\cite{chen2023text2tex} for the horse category (see \cref{fig:rand-poses-grid} for examples rendered using our pipeline), and rely on two hand-crafted textures for the cows and one for the sheep.

\noindent{\bf Iterative refinement and compute.} 
Each refinement iteration of our approach produces completely independent training datasets; we strictly discard all synthetic images and paired 2D-3D instances generated in previous stages

We typically run our refinement process for 4 iterations. Empirical observations (see \cref{fig:ablation-iter}) indicate that the first two iterations yield the most crucial performance gains. The benefit of subsequent iterations is heavily dependent on the natural deformation complexity of the target category: performance on the \textit{horse} category continues to improve up to iteration 7, whereas the \textit{cow} and \textit{sheep} categories generally converge after 3 to 4 iterations (when training on synthetic images). 

Each complete refinement iteration consists of four sequential stages: (1) pseudo-ground-truth inference using the model from the previous iteration ($2-3$ hours), (2) Blender rendering of the new synthetic images ($\sim2$ hours), (3) extraction of DINOv2 and Stable Diffusion features for the training images ($4-5$ hours), and (4) predictor training with gradient accumulation for 60K steps ($\sim 8-10$ hours). This totals approximately 16-20 hours per iteration. Consequently, a 4-iteration model requires roughly 3 days to train. All reported runtimes were measured using 4 NVIDIA RTX A6000 GPUs. Note, that the time required for the first three steps scales with the dataset size. The provided timings correspond to the ca. $16,000$-sample dataset utilized in our ablations.